\documentclass{article}

\newif\ifappendix
\appendixtrue % Comment to remove appendix

\usepackage{amsmath,amsthm,amssymb}
\usepackage{mathtools}
\usepackage{bm}
\usepackage{dsfont}
\usepackage{xspace}
\usepackage{color,soul}
\usepackage{multirow}
\usepackage{multicol}
\usepackage{caption}
\usepackage{enumitem}
\usepackage[linesnumbered,vlined,ruled,noend]{algorithm2e}
\newcommand{\sys}{DFG-NAS\xspace}

\newcommand{\para}[1]{{\vspace{2pt} \bf \noindent #1 \hspace{1pt}}}

\usepackage{CJKutf8}

\newtheorem*{assumption*}{\assumptionnumber}
\providecommand{\assumptionnumber}{}


\newtheorem*{remark*}{\remarknumber}
\providecommand{\remarknumber}{}


\usepackage{microtype}
\usepackage{graphicx}
\usepackage{subfigure}
\usepackage{booktabs} % for professional tables

\usepackage{hyperref}

\usepackage{algorithm}
\usepackage{algorithmic}
\renewcommand{\algorithmicrequire}{\textbf{Input:}}
\renewcommand{\algorithmicensure}{\textbf{Output:}}

\usepackage[accepted]{icml2022}

\icmltitlerunning{DFG-NAS: Deep and Flexible Graph Neural Architecture Search (with appendix)}
\begin{document}

\twocolumn[
\icmltitle{DFG-NAS: Deep and Flexible Graph Neural Architecture Search}
% \ifappendix
% \icmltitle{Don't Waste Your Bits! Squeeze Activations and Gradients for\\Deep Neural Networks via \alg (with appendix)}
% \else
% \icmltitle{Don't Waste Your Bits! Squeeze Activations and Gradients for\\Deep Neural Networks via \alg}
% \fi

% It is OKAY to include author information, even for blind
% submissions: the style file will automatically remove it for you
% unless you've provided the [accepted] option to the icml2020
% package.

% List of affiliations: The first argument should be a (short)
% identifier you will use later to specify author affiliations
% Academic affiliations should list Department, University, City, Region, Country
% Industry affiliations should list Company, City, Region, Country

% You can specify symbols, otherwise they are numbered in order.
% Ideally, you should not use this facility. Affiliations will be numbered
% in order of appearance and this is the preferred way.
\icmlsetsymbol{equal}{*}

\begin{icmlauthorlist}
\icmlauthor{Wentao Zhang}{pku}
\icmlauthor{Zheyu Lin$^{*}$}{pku}
\icmlauthor{Yu Shen$^{*}$}{pku}
\icmlauthor{Yang Li}{pku}
\icmlauthor{Zhi Yang}{pku}
\icmlauthor{Bin Cui}{pku,qd}
\end{icmlauthorlist}

\icmlaffiliation{pku}{School of CS \& Key Laboratory of High Confidence Software Technologies, Peking University}
\icmlaffiliation{qd}{Institute of Computational Social Science, Peking University (Qingdao), China}

\icmlcorrespondingauthor{Zhi Yang}{yangzhi@pku.edu.cn}
\icmlcorrespondingauthor{Bin Cui}{bin.cui@pku.edu.cn}
% You may provide any keywords that you
% find helpful for describing your paper; these are used to populate
% the "keywords" metadata in the PDF but will not be shown in the document
\icmlkeywords{Machine Learning, ICML}

\vskip 0.3in
]

% this must go after the closing bracket ] following \twocolumn[ ...

% This command actually creates the footnote in the first column
% listing the affiliations and the copyright notice.
% The command takes one argument, which is text to display at the start of the footnote.
% The \icmlEqualContribution command is standard text for equal contribution.
% Remove it (just {}) if you do not need this facility.

% \printAffiliationsAndNotice{}  % leave blank if no need to mention equal contribution
\printAffiliationsAndNotice{\icmlEqualContribution} % otherwise use the standard text.

\begin{abstract}
Graph neural networks (GNNs) have been intensively applied to various graph-based applications. 
Despite their success, manually designing the well-behaved GNNs requires immense human expertise.
And thus it is inefficient to discover the potentially optimal data-specific GNN architecture.
This paper proposes DFG-NAS, a new neural architecture search (NAS) method that enables the automatic search of very deep and flexible GNN architectures. Unlike most existing methods that focus on micro-architectures, 
DFG-NAS highlights another level of design: the search for macro-architectures on how atomic propagation (\textbf{\texttt{P}}) and transformation (\textbf{\texttt{T}}) operations are integrated and organized into a GNN. 
To this end, DFG-NAS proposes a novel search space for \textbf{\texttt{P-T}} permutations and combinations based on message-passing dis-aggregation, defines four custom-designed macro-architecture mutations, and employs the evolutionary algorithm to conduct an efficient and effective search. 
Empirical studies on four node classification tasks demonstrate that DFG-NAS outperforms state-of-the-art manual designs and NAS methods of GNNs.
% Graph neural networks (GNNs) have been intensively applied to various graph-based applications. Despite their success, designing good GNN architectures is non-trivial, which heavily relies on lots of human efforts and domain knowledge. Although several attempts have been made in graph neural architecture search, they suffer from the following limitations: 1) fixed pipeline pattern of propagation (P) and (T) transformation operations; 2) restricted pipeline depth of GNN architectures. This paper proposes DFG-NAS, a novel method that searches for deep and flexible GNN architectures. Unlike most existing methods that focus on micro-architecture, DFG-NAS highlights another level of design: the search for macro-architectures of how atomic P and T are integrated and organized into a GNN. Concretely, DFG-NAS proposes a novel-designed search space for the P-T permutations and combinations based on the message-passing dis-aggregation, and defines various mutation strategies and employs the evolutionary algorithm to conduct an efficient and effective search. Empirical studies on four benchmark datasets demonstrate that DFG-NAS could find more powerful architectures than state-of-the-art manual designs and meanwhile are more efficient than the current graph neural architecture search approaches.
\end{abstract}

\section{Introduction}
\label{sec:introduction}
% \zwt{1.GNN应用广泛}
Graph Neural Networks (GNNs) are a set of message passing algorithms, whose intuition is to smooth the node embedding across the edges of a graph. By staking multiple GNN layers, each node can enhance its node embedding with distant neighborhood nodes.
Recently, GNNs have been applied in various domains such as social network analysis~\cite{zhang2020reliable, huang2021knowledge}, chemistry and biology~\cite{DBLP:conf/nips/DaiLCDS19, DBLP:conf/iclr/BradshawKPSH19}, recommendation~\cite{jiang2022zoomer, wu2020graph}, natural language processing~\cite{wu2021graph, vashishth2020graph}, and computer vision~\cite{shi2019skeleton, sarlin2020superglue}. 
%intensively \blue{studied }in recent years.
Despite a broad spectrum of applications, designing GNN architectures manually is a knowledge-intensive and labor-intensive process.
% , which requires immense human expertise and lots of labors to obtain promising performance.
% \blue{As it has been widely observed that handcrafted architectures, such as GCN~\cite{DBLP:conf/iclr/KipfW17}, GraphSAGE~\cite{hamilton2017inductive}, and GAT~\cite{DBLP:conf/iclr/VelickovicCCRLB18}, cannot always behave well in different scenarios and datasets~\cite{huan2021search}, there is an ever-increasing demand for automated architecture exploration to obtain optimal data-specific GNN architectures.}
Existing literature~\cite{huan2021search} suggests that handcrafted architectures (e.g., GCN~\cite{DBLP:conf/iclr/KipfW17}, GraphSAGE~\cite{hamilton2017inductive}, and GAT~\cite{DBLP:conf/iclr/VelickovicCCRLB18}) can not behave well in all scenarios.
Therefore, there is an ever-increasing demand for automated architecture exploration to obtain the optimal data-specific GNN architectures.

Motivated by the success of neural architecture search (NAS) in other established areas, e.g., computer vision, several recent graph neural architecture search (G-NAS) methods are proposed to effectively tackle the architecture challenge in GNNs, including GraphNAS~\cite{gao2019graphnas}, Auto-GNN~\cite{zhou2019auto}, and GraphGym~\cite{you2020design}. 
These G-NAS methods assume GNNs consist of several repetitive message passing layers, and focus more on the intra-layer design such as aggregation function and nonlinear activation function. Despite their effectiveness, they suffer from two limitations:

%GNNs iteratively learns the node embedding via recursive neighborhood propagation ($\texttt{P}$) and transformation ($\textbf{\texttt{T}}$). The learned node embedding could be employed to the downstream graph-based tasks, including  

% \zwt{2.手工设计GNN很难，引出G-NAS}
%To better adapt to different application scenarios, GNNs have been rapidly developed and evolved, leading to a growing number of novel architectures, including GCN~\cite{DBLP:conf/iclr/KipfW17}, GraphSAGE~\cite{hamilton2017inductive}, and GAT~\cite{DBLP:conf/iclr/VelickovicCCRLB18}. 
%Despite the success of GNN models, the existing literature heavily relies on manual architecture design to achieve the best performance, and they are defined and evaluated as specific architectural designs. 

\textbf{Fixed Pipeline Pattern.} To generate GNNs, existing methods adopt a fixed message-passing pipeline to organize two types of atomic operations: 
propagating ($\textbf{\texttt{P}}$) representations of its neighbors and applying
transformation ($\textbf{\texttt{T}}$) on the representations. In particular, most G-NAS methods adopt the tight entanglement of applying
transformation after propagation in each layer (e.g., $\textbf{\texttt{P-T-P-T}}$). Several handcrafted architectures include a certain degree of entanglement by only retaining the first transformation or propagation, e.g., $\textbf{\texttt{T-P-P-P}}$~\cite{DBLP:conf/iclr/KlicperaBG19,liu2020towards} or $\textbf{\texttt{P-P-P-T}}$~~\cite{wu2019simplifying,rossi2020sign, zhang2021graph}. 
However, these specific $\textbf{\texttt{P-T}}$ permutations and combinations are still fixed pipeline designs, limiting the expressive power of macro-architecture search space. It remains to be seen whether more general and flexible pipelines can further improve the performance.

\textbf{Restricted Pipeline Depth.} 
A common practice to increase the expressive power is to directly stack multiple GNN layers~\cite{DBLP:conf/iclr/KipfW17}.
%By stacking multiple GNN layers, one node can propagate its information to distant nodes, and thus improve the node embedding.
%However, the node information may be propagated to the full graph, and 
% But the performance decreases when going deeper} as the node embedding becomes indistinguishable with too many $\texttt{P}$ operations in deep GNNs, leading to the \textit{over-smoothing issue}~\cite{li2018deeper}.
However, when the layers become deeper, the performance decreases as the node embedding becomes indistinguishable with too many \textbf{\texttt{P}} operations, which we refer to as \textit{the over-smoothing issue}~\cite{li2018deeper, miao2021lasagne}.
Therefore, the existing G-NAS methods fix the number of layers to a small constant.
For example, both AutoGNN and GraphNAS pre-define a very restricted GNN layer number (e.g., $\leq 3$). 
How to increase the pipeline depth is another critical problem to conduct effective G-NAS.

This paper proposes DFG-NAS, a new method for automatic search of deep and flexible GNN architectures. Our key insight is that propagation $\textbf{\texttt{P}}$ and transformation $\textbf{\texttt{T}}$ operation  correspond to enforcing and mitigating the effect of smoothing. Inspired by this insight, we propose to search for the best $\textbf{\texttt{P-T}}$ permutations and combinations (i.e., pipeline) to tune suitable 
smoothness level and thus obtain well-behaved data-specific GNN architectures.
Instead of the micro-architecture in intra-layer design, we explore another level of design---\emph{pipeline search} in the GNN macro-architecture.

Specifically, DFG-NAS greatly increases the capabilities of
GNN macro-architecture search in two dimensions: the pipeline pattern and depth of GNN generation.
We accomplish this by dis-aggregate the $\textbf{\texttt{P}}$ and $\textbf{\texttt{T}}$ operations in our search space and define four effective mutation designs to explore $\textbf{\texttt{P-T}}$ permutations and combinations.
To solve the over-smoothing problem, we propose a \textit{gating mechanism} between different $\textbf{\texttt{P}}$ operations so that node-adaptive propagation can be achieved. Furthermore, the skip-connection mechanism is used to $\textbf{\texttt{T}}$ operations to avoid model degradation~\cite{he2016deep, zhang2021evaluating}. Besides the pipeline and mutation designs, an evolutionary algorithm is applied to search for well-behaved GNN architectures.

% \zwt{skip和dense connection选一个，我们加斜体的}
The contribution of the paper is summarized as follows: (1) By decoupling the $\textbf{\texttt{P}}$ and $\textbf{\texttt{T}}$ operations, \sys suggests a transition from studying specific fixed GNN pipelines to
studying the GNN pipeline design space.
% which opens up more opportunities for data-specific GNN architecture innovation.
(2) By further adding gating and skip-connection mechanisms, \sys could support both deep propagation and transformation, which has the ability to explore the best architecture design to push forward the GNN performance boundary.
(3) We search for the flexible pipeline using a custom-designed genetic algorithm so that the final searched GNN architecture represents
the result of joint optimization over the pattern and depth of pipelines.
(4) %We evaluate the performance of \sys on node classification using four public graph datasets.
Empirical results demonstrate that \sys achieves an accuracy improvement of up to 0.9\% over state-of-the-art manual designs and brings up to 15.96x speedups over existing G-NAS methods.

% \zwt{5. Lack of Flexibility, 没有把PT拆开。其他工作search for the optimal combination of operations in a fixed macro-architecture，我们有一个novel perspective，直接从结构入手，研究什么样的宏观结构能提点。}
% GraphGym attempts to search for a simple macro-architecture together with micro-architecture. However, it still suffers from the xxx limitation introduced in Section x.

% \zwt{6.介绍我们的方法}

% \zwt{7.总结contribution}
% \vspace{-4mm}
\section{Preliminary}
\subsection{Problem Formulation}
Given a graph $\mathcal{G}$ = ($\mathcal{V}$, $\mathcal{E}$) with $|\mathcal{V}| = N$ nodes and $|\mathcal{E}| = M$ edges, feature matrix $\mathbf{X} = \{\boldsymbol{x_1},\boldsymbol{x_2} ..., \boldsymbol{x_N}\}$ in which $\boldsymbol{x_i}\in\mathbb{R}^{d}$ is the feature vector of node $v_i$, the node set $\mathcal{V}$ is partitioned into training set $\mathcal{V}_{train}$ (including both the labeled set $\mathcal{V}_l$ and unlabeled set $\mathcal{V}_u$), validation set $\mathcal{V}_{val}$ and test set $\mathcal{V}_{test}$.
Suppose $c$ is the number of label classes, the one-hot vector $\boldsymbol{y}_i\in\mathbb{R}^{c}$ is the ground-truth label for node $v_i$, 
$M$ is the performance evaluation metric of a design in any given graph analysis task, e.g., F1 score or accuracy in the node classification task. 
Specifically, let $\mathcal{F}$ be the search space (finite or infinite) of graph neural architecture.
Graph neural architecture search (G-NAS) aims to find the optimal design $f \in \mathcal{F}$, so that the model can be trained to achieve the best performance in terms of the evaluation metric $M$ on the validation set. Formally, it can be defined as the following bi-level optimization problem:
\begin{equation}
\small
\begin{aligned}
&\mathop{\arg\max}_{f \in \mathcal{F}}  \mathbb{E}_{v_i\in \mathcal{V}_{val}}\left[M\left(\boldsymbol{y}_i, P(\hat{\boldsymbol{y}}_i|f(\theta^{*}))\right)\right],\\
&\mathbf{s.t.}\ : \theta^{*} = \mathop{\arg\min}_{\theta}\ell(f(\theta^{*}), \mathcal{V}_{train}),
\label{eq:target}
\end{aligned}
\end{equation}
\noindent where $P(\hat{\boldsymbol{y}}_i|f(\theta^{*}))$ is the predicted label distribution of node $v_i$, $\ell$ is the loss function and $\theta^{*}$ is the optimized weights of model design $f$. 
For each design $f$, G-NAS first trains the corresponding model weight $\theta$ on the training set $\mathcal{V}_{train}$. Then, it evaluates the trained model $f(\theta^{*})$ on the validation set $\mathcal{V}_{val}$ to obtain the final evaluation result.

\subsection{Graph Neural Networks}
% \zwt{画GCN里的图，P和T都用小圆圈表示}

% \zwt{参考PasCa，以及huxia定义的space，还是按照PT一起来}

Based on the intuitive assumption that locally connected nodes are likely to have the same label, most GNNs iteratively propagate the information of each node to its adjacent nodes, and then transform the information with non-linear transformation. 
We refer to the propagation and transformation operations as $\textbf{\texttt{P}}$ and $\textbf{\texttt{T}}$, respectively.  
At timestep $t$, a message vector $\mathbf{m}^{t}_v$ for node $v\in \mathcal{V}$ is computed with the representations of its neighbors $\mathcal{N}_v$ using the $\textbf{\texttt{P}}$ operation, which is 
$\mathbf{m}^t_v \gets \textbf{\texttt{P}}\left(\left\{\mathbf{h}^{t-1}_u|{u \in \mathcal{N}_v}\right\}\right)$.
Then, $\mathbf{m}^{t}_v$ is then updated according to $\mathbf{h}^t_v \gets \textbf{\texttt{T}}(\mathbf{m}^t_v)$ via the $\textbf{\texttt{T}}$ operation, where \textbf{\texttt{T}} is usually a dense layer. 

Take the vanilla GCN~\cite{DBLP:conf/iclr/KipfW17} as an example, a message passing layer can be formulated as:
\begin{equation}
\small
\begin{aligned}
& \textbf{\texttt{P}}: \quad \quad \mathbf{M}^t = \hat{\mathbf{A}}\mathbf{H}^{t-1},\\
& \textbf{\texttt{T}}: \quad \quad \mathbf{H}^t = \delta(\mathbf{M}^{t}\mathbf{W}^{t}),
\end{aligned}
\label{eq:mp}
\end{equation}
where $\hat{\mathbf{A}}$ is the normalized adjacent matrix, $\mathbf{W}^{t}$ is the training parameters of the $t$-th layer, $\mathbf{M}^t$ and $\mathbf{H}^t$ are matrices formed by $\mathbf{m}^t_v$ and $\mathbf{h}^t_v$, respectively.
By stacking $k$ layers, each node in GCN can utilize the information from its $k$-hop neighborhood. Therefore, the model performance is expected to be improved when more distant neighborhood nodes get involved in the training process.

% The node embedding is recursively propagated and transformed by the $\textbf{\texttt{P}}$ and $\textbf{\texttt{T}}$ operations, respectively.

% A messsage passing layer in GNNs can be formulated as:
% \begin{equation}
% \begin{aligned}
% & \mathbf{m}^t_v \gets \textbf{\texttt{P}}\left(\left\{\mathbf{h}^{t-1}_u|{u \in \mathcal{N}_v}\right\}\right),\\
% &\mathbf{h}^t_v \gets \textbf{\texttt{T}}(\mathbf{m}^t_v).
% \end{aligned}
% \label{eq:mp}
% \end{equation}

% Taking the vanilla GCN as an example, $\textbf{\texttt{P}}$ refers to 
% $\mathbf{M}^t = \hat{\mathbf{A}}\mathbf{H}^{t-1}$, and $\textbf{\texttt{T}}$ refers to $\mathbf{H}^t = \delta(\mathbf{M}^t\mathbf{W}^{(t)})$.

% \subsection{Disentanglement GNNs} 

\subsection{Graph Neural Architecture Search} 
% \zwt{GNAS目标是核心设计P和T operation本身，以及组合形式，which is 呼应到micro和macro}
% \zwt{从Micro和Macro两个角度分析} \zwt{参考huxia写的那篇search space定义}
As the basic operations of GNNs, the search of $\textbf{\texttt{P}}$ and $\textbf{\texttt{T}}$ operations has been widely discussed.
Specifically, the existing G-NAS methods can be classified into micro-architecture search and macro-architecture search. 

\textbf{Micro-architecture search.} Micro-architecture corresponds to the design of graph convolution layers, especially the details for $\textbf{\texttt{P}}$ and $\textbf{\texttt{T}}$ operations. Specifically, the $\textbf{\texttt{P}}$ operation determines how to exchange messages among different nodes in each graph convolution layer. Many GNNs~\citep{DBLP:conf/iclr/KipfW17,wu2019simplifying,DBLP:conf/iclr/KlicperaBG19} adopt the normalized adjacency matrix for neighbor propagation, and several attention mechanisms~\citep{DBLP:conf/iclr/VelickovicCCRLB18, DBLP:conf/ijcai/WangY0L21} are also proposed for more effective propagation. 
In addition, how to combine the propagated node embeddings is also important in $\textbf{\texttt{P}}$ (e.g., $\texttt{MEAN}$,  $\texttt{MAX}$, etc). 
Unlike $\textbf{\texttt{P}}$, which is closely related to the graph structure, $\textbf{\texttt{T}}$ focuses on the non-linear transformation in the neural network, including the choices of hidden size and activation function.
Several G-NAS frameworks~\citep{gao2020graph,zhou2019auto,cai2021rethinking, li2021one} have been proposed to search for the best $\textbf{\texttt{P}}$ and $\textbf{\texttt{T}}$ operations from a pool of various implementations. 
% \red{Unlike these previous works that search for the micro-architecture, our method only focuses on the macro-architecture.}
Different from these works, we focus on macro-architecture.

\textbf{Macro-architecture search.} 
Different from the micro-architecture that highlights the details in each GNN layer, the macro-architecture search focuses on the interaction between different layers. 
Concretely, macro-architecture search involves the design of network topology, including the choices of layer depth and the inter-layer skip connections.
Similar to residual connections and dense connections in CNNs~\citep{he2016deep, huang2017densely}, node representations in one GNN layer do not necessarily solely depend on the immediate previous layer~\citep{DBLP:conf/icml/XuLTSKJ18,li2019deepgcns, miao2021degnn}. 
Take the representative GraphGym~\cite{you2020design} as an example, the search space for macro-architecture includes the choice of graph convolutional layer depth, the pre-processing and post-processing layer depth, and the skip-connections. 
The main difference between our method and previous work is that we open up new design opportunities for macro-architecture search in GNNs.
Specifically, we disentangle the $\textbf{\texttt{P}}$ and $\textbf{\texttt{T}}$ operations in GNN layers,  and thus allow exploring $\textbf{\texttt{P-T}}$ permutations and combinations in designing GNN architectures.
% \zwt{open up 了一个新的design space for Macro}

\section{How do $\textbf{\texttt{P}}$ and $\textbf{\texttt{T}}$ influence GNNs?}

\subsection{Entanglement of GNNs}
% \zwt{介绍decoupled GNN，并基于P和T对现有GNN做分类，参考之前做深的工作}

\begin{figure}[tbp]
	\centering
	\includegraphics[width=0.85\linewidth]{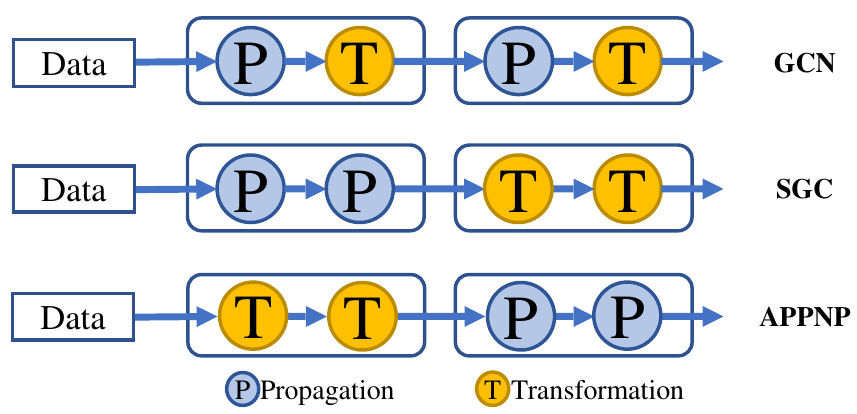}
% 	\vspace{-3mm}
	\caption{\small Entanglement of GNNs.}
	\label{fig:GCN}
% 	\vspace{-5mm}
\end{figure}

\textbf{Entangled GNNs.} The pattern of Entangled Propagation and Transformation is widely adopted by mainstream GNNs, e.g., GCN~\cite{DBLP:conf/iclr/KipfW17}, GraphSAGE~\cite{hamilton2017inductive}, GAT~\cite{DBLP:conf/iclr/VelickovicCCRLB18}, GraphSAINT~\cite{DBLP:conf/iclr/ZengZSKP20}.
Similar to GCN shown in Figure~\ref{fig:GCN}, entangled GNNs pass the input signals through a set of filters to propagate the information, which is further followed by a non-linear transformation.
The propagation operation $\textbf{\texttt{P}}$ and transformation operation $\textbf{\texttt{T}}$ are intertwined and executed alternately in entangled GNNs. 
Therefore, the entangled GNNs share a strict restriction that $D_p = D_t$, where $D_p$ and $D_t$ are the number of propagation and transformation operations, respectively. 

\textbf{Disentangled GNNs.} 
Recently, some researches show that the entanglement of $\textbf{\texttt{P}}$ and $\textbf{\texttt{T}}$ could compromise performance on a range of benchmark tasks~\cite{wu2019simplifying, DBLP:conf/sigir/0001DWLZ020, liu2020towards, sign_icml_grl2020,DBLP:conf/iclr/KlicperaBG19}. They also argue that the true effectiveness of GNNs lies in the propagation operation $\textbf{\texttt{P}}$ rather than the $\textbf{\texttt{T}}$ operation inside the graph convolution. In this way, some disentangled GNNs are proposed to separate $\textbf{\texttt{P}}$ and $\textbf{\texttt{T}}$. 
Following SGC~\cite{wu2019simplifying} as shown in Figure~\ref{fig:GCN}, several methods~\cite{DBLP:conf/sigir/0001DWLZ020, sign_icml_grl2020} execute $\textbf{\texttt{P}}$ operations in advance, and then feed the propagated features into multiple $\textbf{\texttt{T}}$ operations. 
On the contrary, several methods first transform the node features and then propagate the node information to distant neighbors, e.g., APPNP~\cite{DBLP:conf/iclr/KlicperaBG19}, DAGNN~\cite{liu2020towards}, etc.

\subsection{Pipeline Pattern and Depth}
\label{sec:number_order}
The number of $\textbf{\texttt{P}}$ and $\textbf{\texttt{T}}$ operations plays a very important role in GNN learning.. 
As introduced in previous studies~\cite{wu2019simplifying, liu2020towards, miao2021degnn}, more $\textbf{\texttt{P}}$ operations are required to enhance the information propagation when the edges, labels or features are sparse. 
In addition, it is widely recognized that more training parameters and $\textbf{\texttt{T}}$ operations should be used to increase the expressive power of neural networks when the size of dataset (i.e., the number of nodes in a graph) is large. 

% \zwt{\blue{@Zheyu Done}, Table 显示PPTT，TTPP和PTPT三个在PubMed和Cora上的acc}expressive power 

Besides the number of $\textbf{\texttt{P}}$ and $\textbf{\texttt{T}}$ operations, the pipeline pattern (i.e., permutations and combinations) of $\textbf{\texttt{P}}$ and $\textbf{\texttt{T}}$ operations also matters. 
To verify this, we fix the depths of $\textbf{\texttt{P}}$ and $\textbf{\texttt{T}}$ operations to 2 and only change the pipeline pattern. 
We evaluate the three architectures in Figure~\ref{fig:GCN}, and report their test accuracy on three citation networks. 
The experimental results in Table \ref{PTorder} show that the test accuracy varies a lot in the pipeline pattern. Concretely, the $\texttt{TTPP}$ architecture achieves the best performance on PubMed, while it is less competitive than the $\texttt{PPTT}$ architecture on Cora and Citeseer. 
Since the pipeline pattern highly influences the architecture performance and the optimal pipeline pattern varies across different graph datasets, it is necessary to consider the pipeline pattern given a specific dataset/task.

\subsection{Influence on smoothness}
\label{Sec-smo}
% \blue{Both $\textbf{\texttt{P}}$ and $\textbf{\texttt{T}}$ are two basic operations in GNNs.} 
It is widely recognized that the propagation operation in GNN is a Laplacian smoothing, which may lead to indistinguishable node representations ~\cite{li2018deeper}. 
In other words, the representations for all nodes will converge to the same value when GNNs go deeper, which is also called over-smoothing.
In fact, most entangled GNNs (e.g., GCN and GAT) face the over-smoothing problem and suffer from performance degradation when stacking too many GNN layers. 
To better measure the influence of \textbf{\texttt{P}} and \textbf{\texttt{T}} operations on smoothness, we
% \blue{disentangle the propagation and transformation processes in GNNs,} and then 
analyze the change of smoothness when adding a single $\textbf{\texttt{P}}$ or $\textbf{\texttt{T}}$ operation each time. 

\begin{table}[tp]
\caption{Test accuracy of GNNs with different PT orders.}
% \vspace{-5mm}
% \vspace{-2mm}
\centering
{
\noindent
\renewcommand{\multirowsetup}{\centering}
\resizebox{0.8\linewidth}{!}{
\begin{tabular}{cccc}
\toprule
\textbf{Methods} & \textbf{Cora} & \textbf{Citeseer} & \textbf{PubMed} \\
\midrule 
\textbf{\texttt{PPTT}}& 83.4$\pm$0.3&72.2$\pm$0.4 & 78.5$\pm$0.5\\
\textbf{\texttt{TTPP}}& 82.8$\pm$0.2&71.8$\pm$0.3 & 79.8$\pm$0.3 \\
\textbf{\texttt{PTPT}}& 81.2$\pm$0.6&71.2$\pm$0.4 & 79.1$\pm$0.2\\
\bottomrule
\end{tabular}}}
\label{PTorder}
% \vspace{-3mm}
\end{table}

We measure the smoothness by calculating the average similarity~\cite{liu2020towards} between two different node embeddings after a $\textbf{\texttt{P}}$ or $\textbf{\texttt{T}}$ operation. 
Concretely, the node smoothness is defined as follows,
\begin{equation}
\small
S^t_i = 1- \frac{1}{2N-2}\sum_{j\in \mathcal{V}, j\neq i}\left\|\frac{\mathbf{e}^t_i}{||\mathbf{e}^t_i||} - \frac{\mathbf{e}^t_j }{||\mathbf{e}^t_j||}\right\|,
\end{equation}
where $\mathbf{e}^t_i$ is the $i$-th node embedding of the $t$-th GNN layer, and $S^t_i$ measures the similarity of $\mathbf{e}^t_i$ to the entire graph. Larger $\mathbf{e}^t_i$ means node $i$ faces higher risk of over-smoothing issue, and $S^t_i$ ranges in $[0, 1]$ since we adopt the normalized node embedding to compute their Euclidean distance.
Based on the node smoothness $S^t_i$, we further measure the average similarities between all the node pairs and define the smoothness of the whole graph $\mathcal{G}$ as:
\begin{equation}
\label{equation:smoothness}
\small
S^t = \frac{1}{N}\sum_{i\in \mathcal{V}}S^t_i,
\end{equation}
where $S^t$ is the graph smoothness of the $t$-th GNN layer.

\begin{figure}
	\centering
	\includegraphics[width=0.85\linewidth]{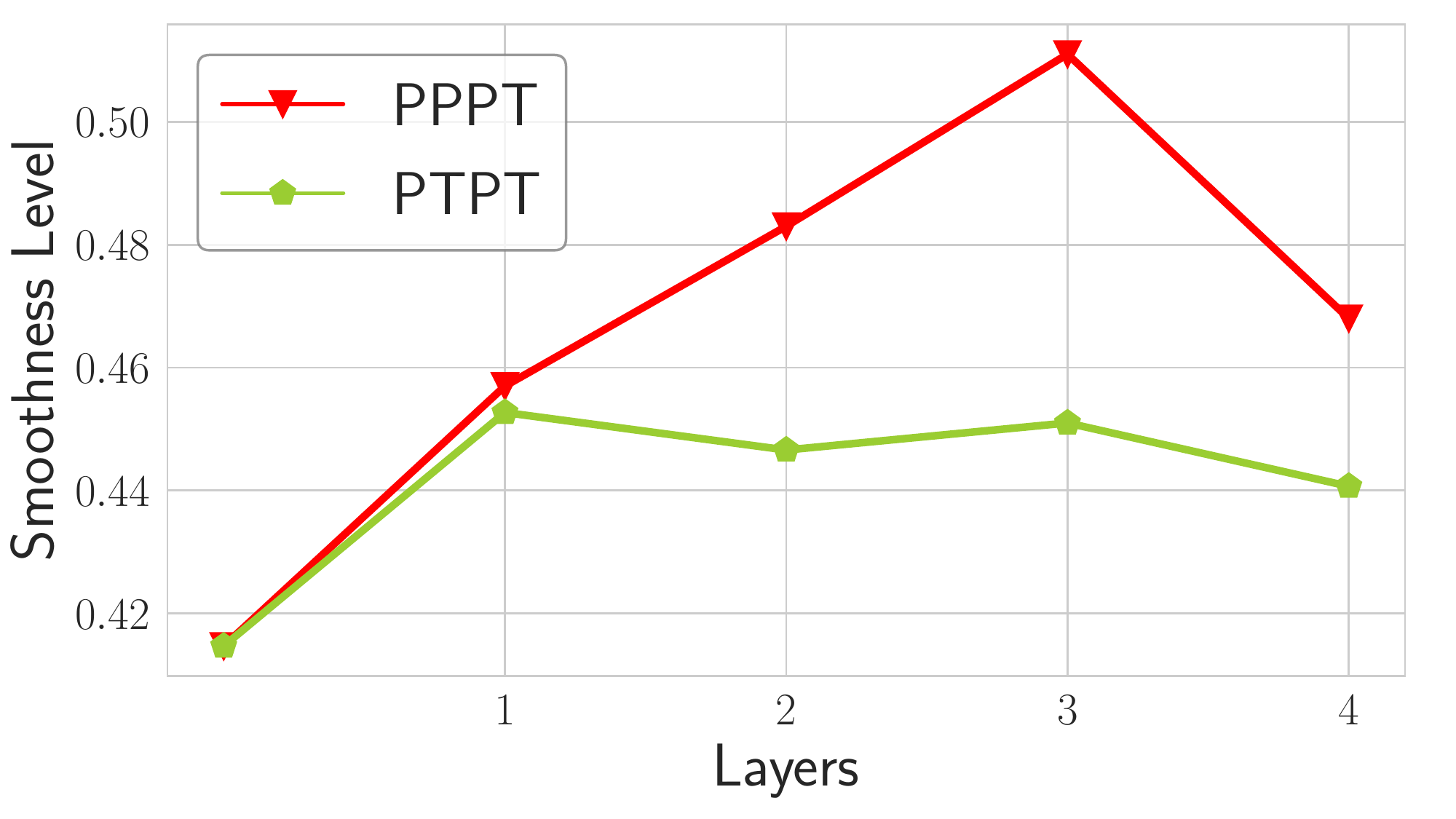}
% 	\vspace{-3mm}
	\caption{\small Smoothness of different \textbf{\texttt{PT}} orders.}
	\label{fig:smoothness}
% 	\vspace{-3mm}
\end{figure}

% \zwt{\blue{@Zheyu Done}, 这里插入两张图，分别是PPPT以及PTPTPT以及对应的smoothness}
Unlike previous G-NAS methods that treat the combination of $\textbf{\texttt{P}}$  $\textbf{\texttt{T}}$ as one GNN layer, we treat each $\textbf{\texttt{P}}$ or $\textbf{\texttt{T}}$ here as an individual layer.
We change different permutations and combinations of $\textbf{\texttt{P}}$ and $\textbf{\texttt{T}}$, and report the corresponding output smoothness of each layer in SGC and GCN on Cora. 
As shown in Figure~\ref{fig:smoothness}, the results show that 1) the smoothness increases, i.e., the node embedding becomes similar, by applying the $\textbf{\texttt{P}}$ operation; 2) the smoothness decreases by applying the $\textbf{\texttt{T}}$ operation, which implies that the $\textbf{\texttt{T}}$ operation has the ability to alleviate the over-smoothing issue. 
While over-smoothing leads to indistinguishable node embedding, under-smoothing cannot unleash the full potential of the graph structure.
Therefore, how to carefully control the smoothness with $\textbf{\texttt{P}}$ or $\textbf{\texttt{T}}$ in GNNs is also an open question.

% \zwt{P smooth, T anti-smooth}

% \zwt{说1）P和T的个数会影响 2）P和T的顺序会影响}

\section{The Proposed Method}
Based on the message passing dis-aggregation, we propose a general design space for GNN pipeline search, which includes the crucial aspects of the number, and the permutations and combinations of $\textbf{\texttt{P}}$ and $\textbf{\texttt{T}}$ operations. Then we provides a custom-designed genetic
algorithm to search the space of pipeline pattern and depth.

%explain how we allow deep propagation and transformation in GNNs. After that, we introduce the search method and discuss the properties of our proposed algorithm.  

% \zwt{参考沈公子发的遗传算法做NAS工作，说这个类似Gene一样，字数不够可以用encoder编码成01来表示}

% \zwt{四个P中间插入一个T，四个T中间插入一个P,也weighted加起来，只是weights写在上面

\subsection{Pipeline Search Space}
% \zwt{说我们空间允许P和T的组合}
Our general search space of GNNs pipeline includes $\textbf{\texttt{P-T}}$ permutations and combinations, and the number of $\textbf{\texttt{P-T}}$ operations. Besides, we further add connections among each type of operation to enable deep propagation and transformation in \sys. 
Suppose a single $\textbf{\texttt{P}}$ or $\textbf{\texttt{T}}$ operation is one GNN layer in \sys, $\mathbf{o}^{(l)}_v$ is the output of node $v$ in the $l$-th layer, $L_P$ and $L_T$ are two sets that include the layer index of all $\textbf{\texttt{P}}$ operations and $\textbf{\texttt{T}}$ operations, respectively. The layer connections between different layers is as follows.

% We treat a single $\textbf{\texttt{P}}$ or $\textbf{\texttt{T}}$ operation as one GNN layer in \sys. Denote $\mathbf{o}^{(l)}_v$ as the output of node $v$ in the $l$-th GNN layer.
% $L_P$ and $L_T$ are two sets that include the layer index of all $\textbf{\texttt{P}}$ operations and $\textbf{\texttt{T}}$ operations, respectively. \blue{Besides the permutations and combinations, we further add connection among each type of operations to} enable deep propagation and transformation in \sys as follows.

\textbf{Propagation Connection.}  
As introduced in Section~\ref{Sec-smo}, GNNs may suffer from the over-smoothing or under-smoothing issue if we execute too many or too few propagation operations. In addition, a previous study~\cite{zhang2021node} shows that the nodes with different local structures have different smoothing speeds. Therefore, how to control the smoothness of different nodes in a node-adaptive way is essential in GNN architecture design.

\begin{figure}[tpb]
	\centering
	\includegraphics[width=0.95\linewidth]{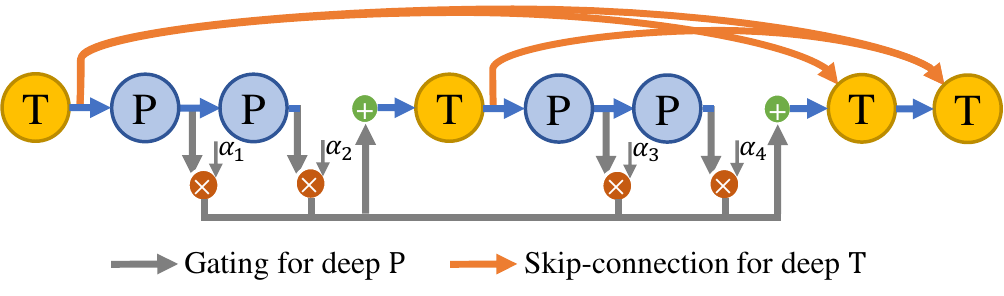}
% 	\vspace{-3mm}
	\caption{\small GNN pipeline example in the search space of \sys.}
	\label{fig:enable}
% 	\vspace{-5mm}
\end{figure}

To allow deep propagation and provide suitable smoothness for different nodes, we adopt a gating mechanism upon $\textbf{\texttt{P}}$ operations. 
The output of the $l$-th $\textbf{\texttt{P}}$ operation is the propagated node embedding of $\mathbf{o}^{(l-1)}$ if its next operation is $\textbf{\texttt{P}}$. 
As shown in Figure~\ref{fig:enable}, if the next operation is $\textbf{\texttt{T}}$, we assign a node-adaptive combination weight for the node embeddings propagated by all previous $\textbf{\texttt{P}}$ operations.
The above process can be formulated as,
% \begin{equation}
% \begin{aligned}
%       & \mathbf{c_{msg}} \gets \sum_{\mathbf{m}_v^i\in M_v} \alpha_i \mathbf{m}_v^i,\  \alpha_i = \text{Softmax}(\sigma(\mathbf{s} \mathbf{m}_v^i)),
% \label{eq:gate}
% \end{aligned}
% \end{equation}
% \begin{equation}
% \small
% \mathbf{o}^{(l)}_v=\left\{
% \begin{aligned}
% &\textbf{\texttt{P}}(\mathbf{o}^{(l-1)}_v),      & \text{Followed by}\ \textbf{\texttt{P}}\\
% &\sum_{i\in L_P, i<l} \alpha_i \mathbf{o}_v^i + \alpha_l \textbf{\texttt{P}}(\mathbf{o}^{(l-1)}_v),      &  \text{Followed by} \  \textbf{\texttt{T}}\\
% \end{aligned}
% \right.,
% \end{equation}
\begin{equation}
\small
\begin{aligned}
\mathbf{z}^{(l)}_v&=\textbf{\texttt{P}}(\mathbf{o}^{(l-1)}_v),\\
\mathbf{o}^{(l)}_v&=\left\{
\begin{aligned}
&\mathbf{z}^{(l)}_v,      & \text{Followed by}\ \textbf{\texttt{P}}\\
&\sum_{i\in L_P, i \leq l} \text{Softmax}(\alpha_i) \mathbf{z}_v^{(i)},      &  \text{Followed by} \  \textbf{\texttt{T}}\\
\end{aligned}
\right.,
\end{aligned}
\end{equation}
where $\alpha_i = \sigma(\mathbf{s} \cdot \mathbf{o}_v^i)$ is the weight for $i$-th layer output of node $v$. 
$\mathbf{s}$ is the trainable vector shared by all nodes , and $\sigma$ denotes the Sigmoid function
We adopt the Softmax function to scale the sum of gating scores to 1.

% \zwt{$\alpha_i$重复定义。第一个P操作会对后续很多层产生gate权重，但各不相同，不能使用同一个字母。可能应该是$\alpha_i^{(l)}=\text{Softmax}(\sigma(\mathbf{s} \cdot \mathbf{o}_v^i))$}
% \zwt{cmsg改成z}
% \zwt{对于P的输出，如果下一层是P那就直接propagate，否则如果是T就要propagate并且之后和前面所有层propagate之后的weighted加起来}
% \zwt{对于gate-connetion而言，我们定义连接规则为对T之前的所有P进行gate连接，以自适应地选取特征，解决over-smoothing问题；
% 对于res-connection，我们定义连接规则为对前后不同的T，设置残差连接(这里我们用的dense连接的方式)，来解决模型退化问题；
% }

\textbf{Transformation Connection.} 
% \zwt{左边序列..PTTTP..，画结构出来, 参考DenseNet}
GNNs require more training parameters and non-linear transformation operations to enhance their expressive power on larger graphs. However, it is widely acknowledged that too many transformation operations will lead to the \textit{model degradation} issue~\cite{he2016deep}, i.e., both the training and test accuracies decrease along with the increased transformation steps.

To allow deeper transformation and meanwhile alleviate the \textit{model degradation} issue, we introduce the skip-connection mechanism into $\textbf{\texttt{T}}$ operations. 
As shown in Figure~\ref{fig:enable}, the input of each $\textbf{\texttt{T}}$ operation is the sum of the output of the last layer and the outputs of all previous $\textbf{\texttt{T}}$ operations before the last layer. 
Concretely, the input and output of the $l$-th $\textbf{\texttt{T}}$ operation can be defined as follows:
\begin{equation}
\begin{aligned}
\small
% \mathbf{h}^{(l+1)}_v = \sigma(\mathbf{h}^{(l)}_v\mathbf{W}^{(l)})+ \sum_{i =1}^{l} \mathbf{h}^{(i)}_v,
% \mathcal{G}_{t+1}=\mathcal{F}\left(\mathcal{G}_{t}, \mathcal{W}_{t}\right)+\mathcal{G}_{t}=\mathcal{G}_{t+1}^{r e s}+\mathcal{G}_{t}
\mathbf{z}^{(l)}_v&=\mathbf{o}^{(l-1)}_v + \sum_{i\in L_T, i< m(l)} \mathbf{o}^{(i)}_v, \\
\mathbf{o}^{(l)}_v&=\sigma\left(\textbf{z}^{(l)}_v\mathbf{W}^{(l)}\right),
\end{aligned}
\end{equation}
where $m(l)$ is the index of the last $\textbf{\texttt{T}}$ operation before the $l$-th layer, and
$\mathbf{W}^{(l)}$ is the learnable parameter in the $l$-th $\textbf{\texttt{T}}$ operation.
Specifically, the first layer in our search space is a \textbf{\texttt{T}} operation, and $\mathbf{z}^{(1)}$ refers to the original node features.
% $\mathbf{h}^{(l)}_v$ is the embedding of node $v$ after $l$ times $\textbf{\texttt{T}}$ operation. 
% \zwt{上一层可能是P，那就是不同的P weighted加起来的结果，如果是T可能加俩遍。transformation算输入，propagation可以算输出。分成两行，一行是加起来看input，第二行是sigma激活函数。现在每一层都tensor，不要h和m}
% \zwt{layer index比如所有的P对应的layer，写个set：the set of layer indices}
% \zwt{z是输入，o是输出}

% \zwt{最后检查全文search space和search space的混用}

\subsection{Search via Genetic Algorithm}
Genetic algorithm (GA) aims to evolve and improve an entire population of individuals via nature-inspired mechanisms such as mutations.
In \sys, an individual’s chromosome represents a GNN pipeline. 
To create a new generation of individuals, we perform
the following mutations on the chromosomes of the current generation.

\begin{figure}[tpb]
	\centering
	\includegraphics[width=0.9\linewidth]{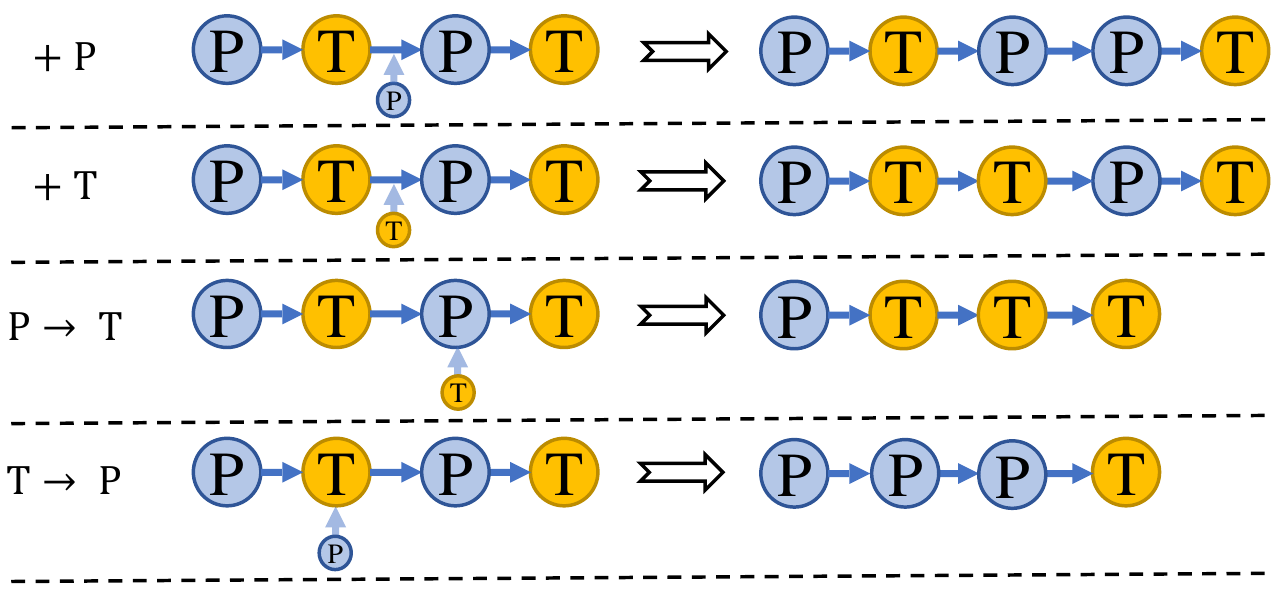}
% 	\vspace{-3mm}
	\caption{Overview of four different mutations.}
	\label{fig:muta}
	\vspace{-5mm}
\end{figure}

\para{Mutation Designs.}
\label{mu-d}Evolutionary algorithms are a class of optimization algorithms inspired by biological evolution. Specifically, they apply mutations on a population of designs, i.e., the set of different GNN architectures. 
In \sys, each GNN architecture is encoded as a sequence consisting of the $\textbf{\texttt{P}}$ and $\textbf{\texttt{T}}$ operations. 
As shown in Figure~\ref{fig:muta}, we design four mutations, which are as follows,

\textbf{Case 1 + $\textbf{\texttt{P}}$:} Add a propagation operation.

\textbf{Case 2 + $\textbf{\texttt{T}}$:} Add a transformation operation.

\textbf{Case 3 $\textbf{\texttt{P}} $ $\rightarrow$ $\textbf{\texttt{T}}$:} Replace a propagation operation by a transformation operation.

\textbf{Case 4 $\textbf{\texttt{T}}$ $\rightarrow$ $\textbf{\texttt{P}}$:} Replace a transformation operation by a propagation operation.

The above four mutations take place at a random position of the sequence. For example, we can add a propagation operation after or before any other operations. These four specific mutations are proposed for their similarity to the actions that a human designer may take when improving the GNN architecture.  

\para{Evolutionary Algorithm.}To perform an efficient and effective search on our search space, we adopt the evolutionary algorithm as the searching method, which is a class of optimization algorithms inspired by biological evolution. 
The concrete searching pipeline is summarized in Algorithm~\ref{alg:search}.

We randomly generate $k$ different GNN architectures as initial individuals in a population set $\mathbf{Q}$ (line 1), and then evaluate these GNNs on the validation set (line 2).
Next, we randomly sample $m$ ($m < k$) individuals from the population (line 4) and select the architecture with the best validation performance as parent $A$ (line 5). 
After that, the child GNN architecture $B$ is generated by randomly picking one of the four mutations (i.e., + $\textbf{\texttt{P}}$, + $\textbf{\texttt{T}}$, $\textbf{\texttt{P}}$ $\rightarrow$ $\textbf{\texttt{T}}$, or $\textbf{\texttt{T}}$ $\rightarrow$ $\textbf{\texttt{P}}$) introduced in Section~\ref{mu-d}. 
At last, $B$ is evaluated and added to the population (line 7), and then the oldest individual in $\mathbf{Q}$ is removed (line 8).
The above process is repeated for $T$ generations and finally returns the architecture with the best observed evaluation performance (line 9). 
% \blue{Note that the best observed architectures are recorded while updating the population, and we return the best GNN architecture after sufficient updating steps.}
% \zwt{这是啥步骤，为啥要这个，这不是做实验才关心的东西吗，算法设计里说保存了啥东西干啥，这也不是为了算法运行需要保存的东西}

\begin{algorithm}[t]
  \SetAlgoLined
  \KwIn{The population set $\mathbf{Q}$, the maximum generation times $T$, the generation number $t = 1$.}
  \KwOut{The best GNN architecture.}
  \SetAlgoLined
  \caption{\textbf{Searching method.}}
  \label{alg:search}
   Initialize the population set $\mathbf{Q}$ with $|\mathbf{Q}| = k$;
   
   Evaluating $k$ different GNN architectures in $\mathbf{Q}$;
   
   \For{$1\leq t\leq T$}
   {
       Randomly sample $m$ individuals from $\mathbf{Q}$;
       
       Select the parent $A$ with best evaluation performance from the $m$ individuals;
       
       Randomly mutate $A$ with the mutation designs, and get the mutated individual $B$;
       
       Evaluate $B$ and adding it to $\mathbf{Q}$;
       
       Remove the oldest individual in $\mathbf{Q}$;
   }
   \textbf{return} the individual with best evaluation performance.
\end{algorithm}

% \subsection{Discussions}
% \textbf{Novelty Statement.}
% \zwt{DG-NAS是第一个搜索PT组合的工作，并且可以做深}

% \zwt{给P和T一个definition--崔老师建议}
\subsection{Relationship with Existing Literature}
The proposed \sys differs from existing G-NAS methods in both the pipeline pattern and pipeline depth of the searched architectures.
% \zwt{这里缺少连接，上下两句话逻辑是啥来着, difference不需要引子-挪到前面来，直接说清楚}

\textbf{Fixed vs. Flexible.} The existing G-NAS methods treat the combination of propagation and transformation as one GNN layer, which leads to a fixed pipeline pattern. 
Instead, \sys disentangles the propagation and transformation operations and thus enjoys better flexibility.

\textbf{Shallow vs. Deep.} The over-smoothing issue in deep GNNs has not been well studied and considered in the search space of the existing methods. 
For example, both GraphNAS and AutoGNN adopt a search space that applies a fixed three-layer GNN as the macro-architecture. 
% Although GraphGym enables various choices of GNN depths by introducing skip-connections, the over-smoothing issue limits its ability to explore deeper GNN architectures. 
\sys enables both deep $\textbf{\texttt{P}}$ and $\textbf{\texttt{T}}$ with the \textit{gating mechanism} and \textit{skip-connection mechanism}, respectively.

Note that \sys mainly focuses on 
GNN macro-architecture search, so it is orthogonal to and compatible with the GNN micro-architecture search space (e.g, aggregation/activation functions, batch normalization, etc.) used by existing G-NAS methods.
%the GNN architectures can be further optimized with micro-architecture search introduced by other G-NAS methods.

% \zwt{Overclaim了！ 之前也陈述了GraphGym了有marco的部分，
% 这里要精准体现novelty}

% \zwt{和现有方法比有啥优势：1）做深 2）更灵活，并且可以结合在现有方法上 }
% \zwt{可以和现有方法结合，就算不结合也比他们强}

% \textbf{Time Complexity}
% \zwt{和GraphNAS和AutoGNN比较一下}

\section{Experiments and Results}
\label{exp_sec}
To evaluate \sys, we apply it on four public graph datasets. Compared with the state-of-the-art baselines, we list three main insights that we will investigate as follows,
\begin{itemize}
    \item \sys generates more powerful architectures than state-of-the-art manual designs. 
    \item \sys works more efficiently than other NAS methods for GNN. It reaches similar performance to other methods while spending less search time.
    \item \sys is able to tackle the two limitations as mentioned in Section \ref{sec:introduction}. Concretely, the Gate operation helps avoid over-smoothing while the disentanglement of \textbf{\texttt{P}} and \textbf{\texttt{T}} increases architecture flexibility.
\end{itemize}

\subsection{Experimental Setup}
\para{\textbf{Baselines.}}
We compare our proposed method \sys with nine manual GNN architectures and three NAS methods for GNN. The manual designs include the following three types of GNNs according to their pipeline pattern of propagation (\textbf{\texttt{P}}) and transformation (\textbf{\texttt{T}}) operations.
\begin{itemize}
    \item \textbf{ Alternate \texttt{P} and \texttt{T}:} GCN ~\cite{DBLP:conf/iclr/KipfW17}, GraphSAGE ~\cite{hamilton2017inductive}, and GAT ~\cite{velivckovic2017graph}. The propagation operation \textbf{\texttt{P}} and transformation operation \textbf{\texttt{T}} are intertwined and executed alternately in these entangled GNNs.
    \item \textbf{ \texttt{T} before \texttt{P}:} APPNP~\cite{DBLP:conf/iclr/KlicperaBG19}, AP-GCN~\cite{spinelli2020adaptive}, and DAGNN~\cite{liu2020towards}. These disentangled GNNs firstly transform the node features with \textbf{\texttt{T}} operations, and then propagate the outputs with multiple \textbf{\texttt{P}} operations. 
     \item \textbf{ \texttt{P} before \texttt{T}:} SGC~\cite{wu2019simplifying} , SIGN~\cite{rossi2020sign}, and S$^2$GC~\cite{zhu2021simple}. On contrary to APPNP, these disentangled GNNs execute multiple \textbf{\texttt{P}} operations in advance, and then transform propagated features with \textbf{\texttt{T}} operations. 
\end{itemize}

More details of the above manual methods are provided in Appendix~\ref{baseline_methods}. 
Besides, the compared G-NAS methods include:
(1) Auto-GNN~\cite{zhou2019auto}: a reinforced conservative search strategy by adopting both RNNs and evolutionary algorithms in the controller; 
(2) GraphNAS~\cite{gao2019graphnas}: a reinforcement learning-based method that uses an RNN controller to sample from the multiple architectures sequentially; 
(3) GraphGym~\cite{you2020design}: a variant of random search on a general GNN search space that considers intra-layer design, inter-layer design, and training configurations.

\para{\textbf{Datasets.}}
% To evaluate the performance of \sys and ensure reproducibility, 
We conduct the experiments on four public graph datasets: three citation graphs (Cora, Citeseer and PubMed)~\cite{DBLP:conf/iclr/KipfW17}, and one large OGB graph (ogbn-arxiv)~\cite{DBLP:conf/nips/HuFZDRLCL20}. 
We follow the public training/validation/test split for three citation networks and adopt the official split in the OGB graph. The statistics of these datasets are summarized in Appendix~\ref{dataset_description}.

\para{\textbf{Experimental Settings.}}
For a fair end-to-end comparison on each task, we run \sys for 500 iterations, and set the time cost as the budget for other G-NAS baselines.
Then we report the average best observed test accuracy of the searched architecture. 
Specifically, we re-training each searched architecture ten times to avoid randomness. 
These baselines are implemented based on their open-sourced version.
Since AutoGNN is not publicly available, we only report its performance on citation graphs following its paper.
% Note that the code of GraphNAS and GraphGym is publicly available, and we follow their public repositories in the experiments. 
% Besides, we directly use the reported results of AutoGNN in three citation networks, and ignore the results of Auto-GNN in ogbn-arxiv dataset since this method is not open-sourced. 
The details of hyperparameters and reproduction instructions are provided in Appendix~\ref{hpo} and \ref{eno}.
% \zwt{我们Based on了开源仓库，AutoGNN没实现直接拉的结果}

% \zwt{Cora的数值变高了，实验分析也要改}

\begin{table}[tpb!]
% \vspace{-2mm}
\caption{Test accuracy on the node classification task.}
% \vspace{-2mm}
\centering
{
\noindent
\renewcommand{\multirowsetup}{\centering}
\resizebox{0.95\linewidth}{!}{
\begin{tabular}{cccccc}
\toprule
\textbf{Methods} & \textbf{Cora} & \textbf{Citeseer} & \textbf{PubMed} & \textbf{ogbn-arxiv}\\
\hline
 \multicolumn{5}{c}{\textbf{Alternate \texttt{P} and \texttt{T}}}\\
GCN& 81.3$\pm$0.6 & 71.1$\pm$0.1 &78.8$\pm$0.4 & 71.7$\pm$0.3\\
GAT& 82.9$\pm$0.2 & 70.8$\pm$0.5 &79.1$\pm$0.1 & 71.9$\pm$0.2\\
GraphSAGE& 79.2$\pm$0.6 & 71.6$\pm$0.5 & 77.4$\pm$0.5 & 71.5$\pm$0.3 \\
% JK-Net& 81.8$\pm$0.5  & 70.7$\pm$0.7 & 78.8$\pm$0.7 & 72.2$\pm$0.2 \\
% ResGCN& 81.2$\pm$0.5  & 70.8$\pm$0.4 & 78.6$\pm$0.6 & 72.6$\pm$0.4 \\
\hline
 \multicolumn{5}{c}{\textbf{\texttt{T} before \texttt{P}}}\\
APPNP& 83.1$\pm$0.5 & 71.8$\pm$0.4 & 80.1$\pm$0.2 & 72.0$\pm$0.1 \\
AP-GCN& 83.4$\pm$0.3& 71.3$\pm$0.5& 79.7$\pm$0.3 & 71.9$\pm$0.2 \\
DAGNN & 84.3$\pm$0.2 & 73.3$\pm$0.6 & 80.5$\pm$0.5 & 72.0$\pm$0.3 \\
\hline
 \multicolumn{5}{c}{\textbf{\texttt{P} before \texttt{T}}}\\
SGC & 81.7$\pm$0.2 & 71.3$\pm$0.2 & 78.8$\pm$0.1 & 71.6$\pm$0.3 \\
SIGN & 82.1$\pm$0.3 & 72.4$\pm$0.8 & 79.5$\pm$0.5 & 71.9$\pm$0.1 \\
S$^2$GC& 82.7$\pm$0.3 & 73.0$\pm$0.2 &79.9$\pm$0.3 & 71.8$\pm$0.3 \\
% GBP& 83.9$\pm$0.7 & 72.9$\pm$0.5 &\underline{80.6$\pm$0.4} & 72.2$\pm$0.2 \\
\hline 
\multicolumn{5}{c}{\textbf{G-NAS Methods}}\\
GraphNAS & 83.7$\pm$0.4 & 73.5$\pm$0.3 & 80.5$\pm$0.3 &  71.7$\pm$0.2 \\
AutoGNN & 83.6$\pm$0.3 & 73.8$\pm$0.7 & 79.7$\pm$0.4 & / \\
GraphGym & 83.5$\pm$0.2 & 73.4$\pm$0.3 & 80.3$\pm$0.2 & 71.6$\pm$0.3\\
\hline
\sys & \textbf{85.2$\pm$0.2} & \textbf{74.1$\pm$0.4} & \textbf{81.1$\pm$0.3} & \textbf{72.3$\pm$0.2} \\
\bottomrule
\end{tabular}}}
\label{table:performance}
\vspace{-3mm}
\end{table}

\subsection{Comparison with Existing GNNs}
We first compare the architectures obtained by \sys with state-of-the-art manual designs on four datasets in Table \ref{table:performance}. 
The structures of the searched architectures are provided in Appendix~\ref{sec:searched_results}. 
Among the manual designs, DAGNN achieves stable and competitive performance on four datasets.
The reason is that it applies a single Gate operation to combine all propagation outputs, and in this way, it allows a deeper propagation than other manual designs.
However, DAGNN employs the straight-forward MLP for transformation, which may lead to the model degradation issue when transformation goes deeper.
In addition, as all the compared manual designs follow a specific pipeline pattern of \textbf{\texttt{P}} and \textbf{\texttt{T}}, their performance can not be further improved due to their restricted flexibility.
As a result, \sys obtains the deepest architectures with high expressive power and achieves the best performance on all four datasets.
Remarkably, the architecture searched by \sys on the large-scale ogbn-arxiv dataset contains 27  \textbf{\texttt{P}} and 16 \textbf{\texttt{T}} operations, and it outperforms the best manual design DAGNN by a margin of 0.3\% on test accuracy.

\begin{figure*}
% \vspace{-1mm}
	\centering
	\subfigure[Cora]{
    \label{fig:nas_cora}
		% Requires \usepackage{graphicx}
		\scalebox{0.235}{
    \includegraphics[width=0.98\linewidth]{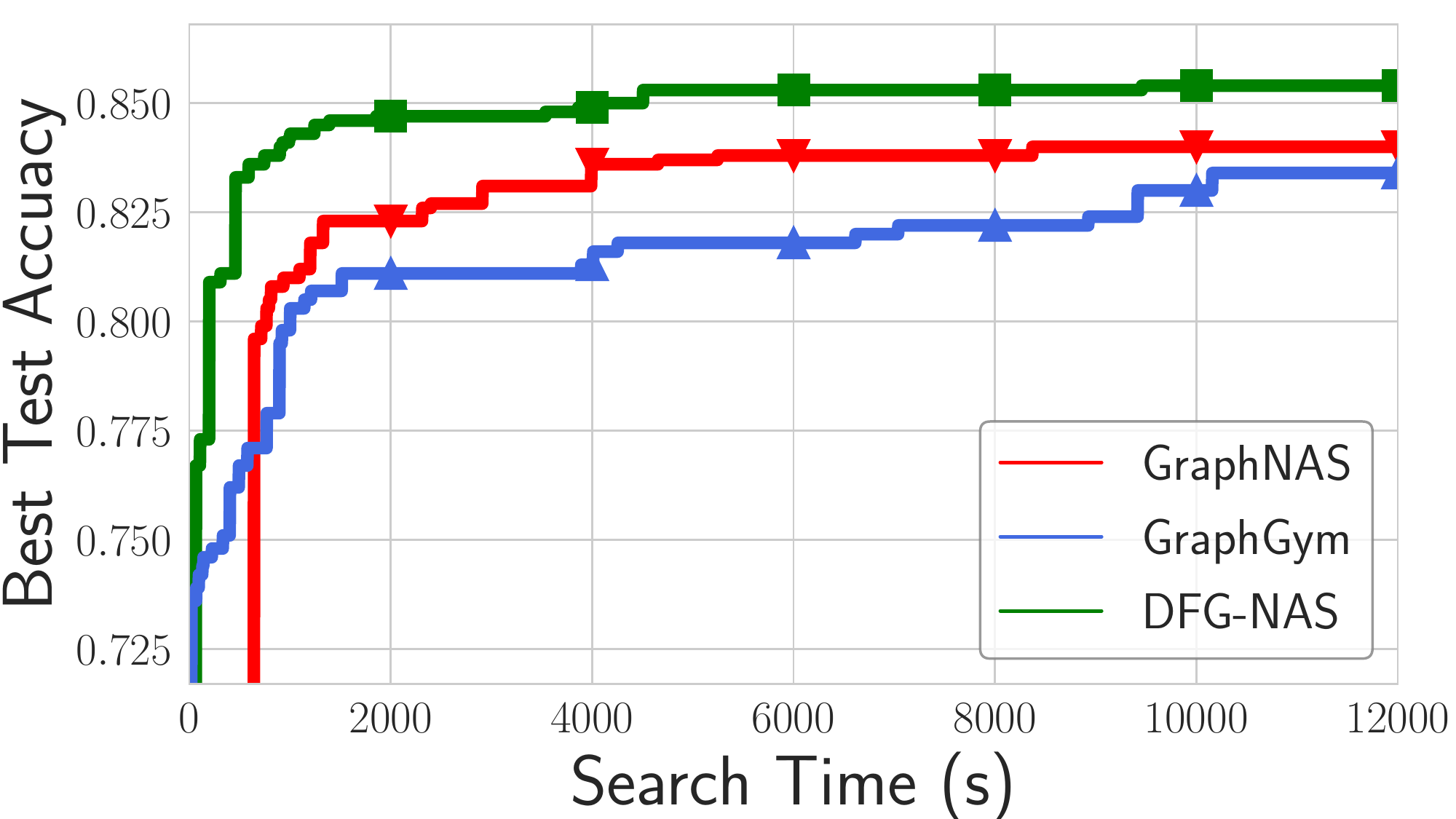}}
	}
	\subfigure[Citeseer]{
    \label{fig:nas_citeceer}
		% Requires \usepackage{graphicx}
		\scalebox{0.235}{
    \includegraphics[width=0.98\linewidth]{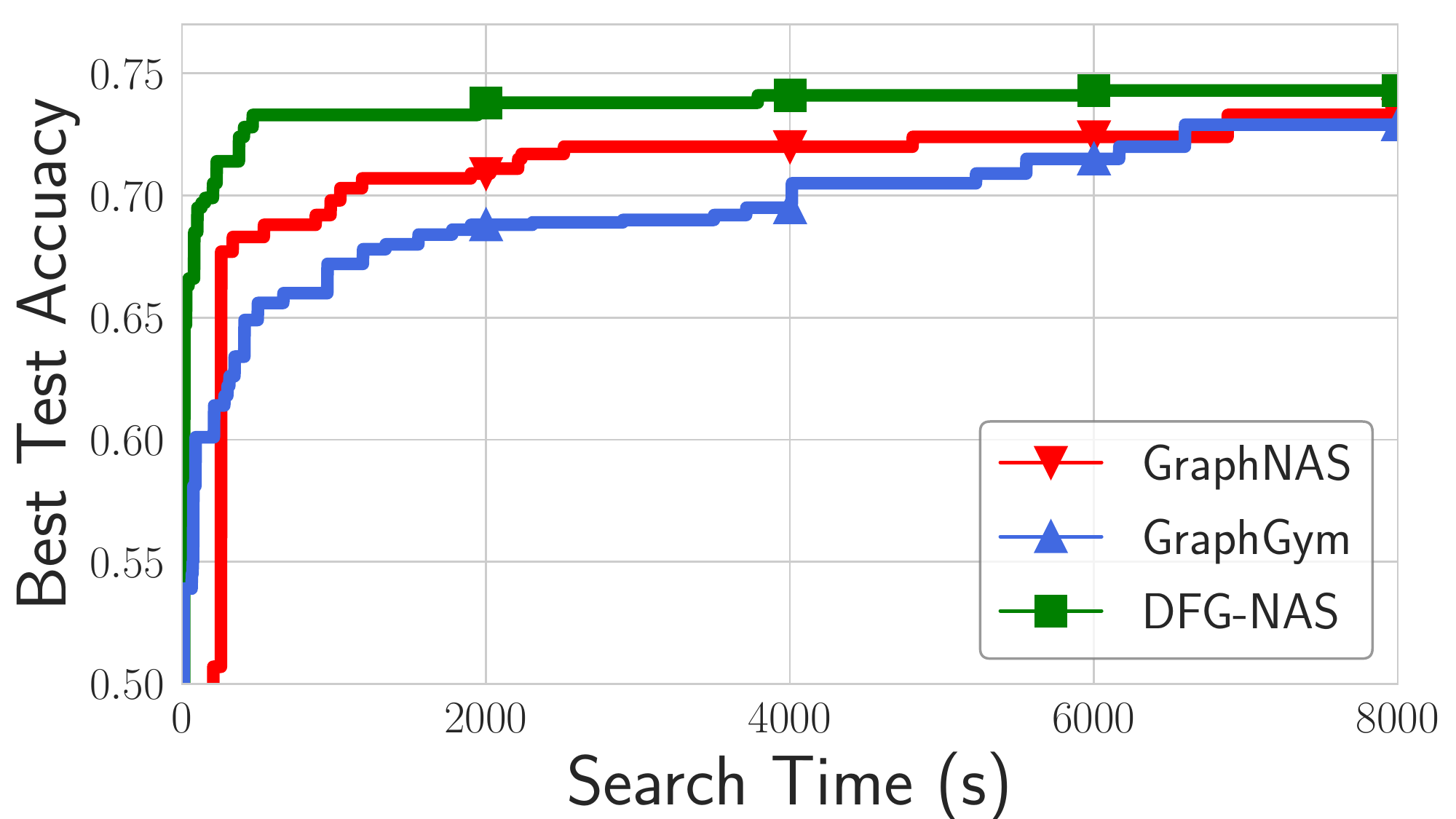}}
	}
	\subfigure[Pubmed]{
    \label{fig:nas_pubmed}
		% Requires \usepackage{graphicx}
		\scalebox{0.235}{
    \includegraphics[width=0.98\linewidth]{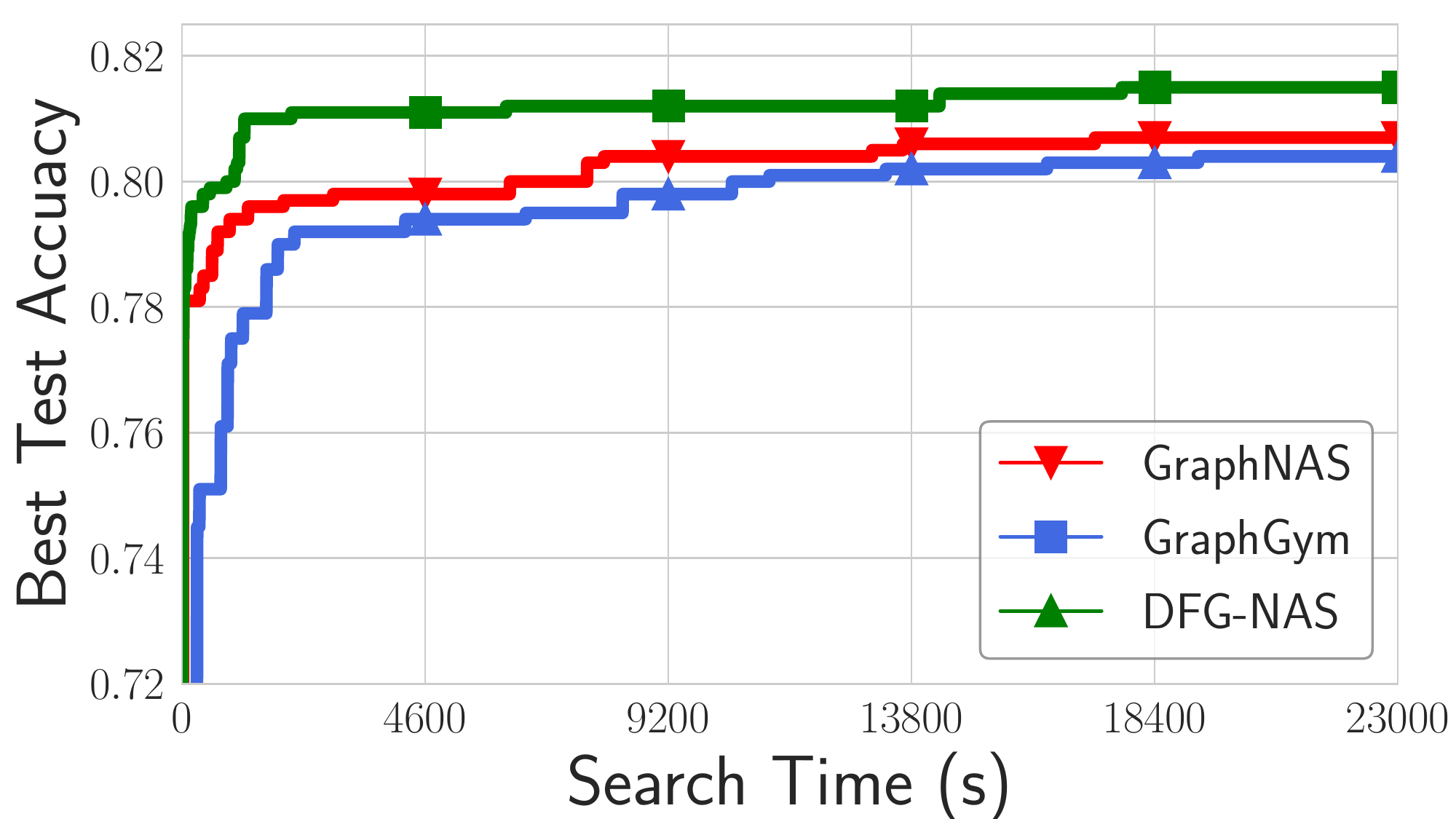}}
	}
	\subfigure[ogbn-arxiv]{
    \label{fig:nas_ogb}
		% Requires \usepackage{graphicx}
		\scalebox{0.235}{
    \includegraphics[width=0.98\linewidth]{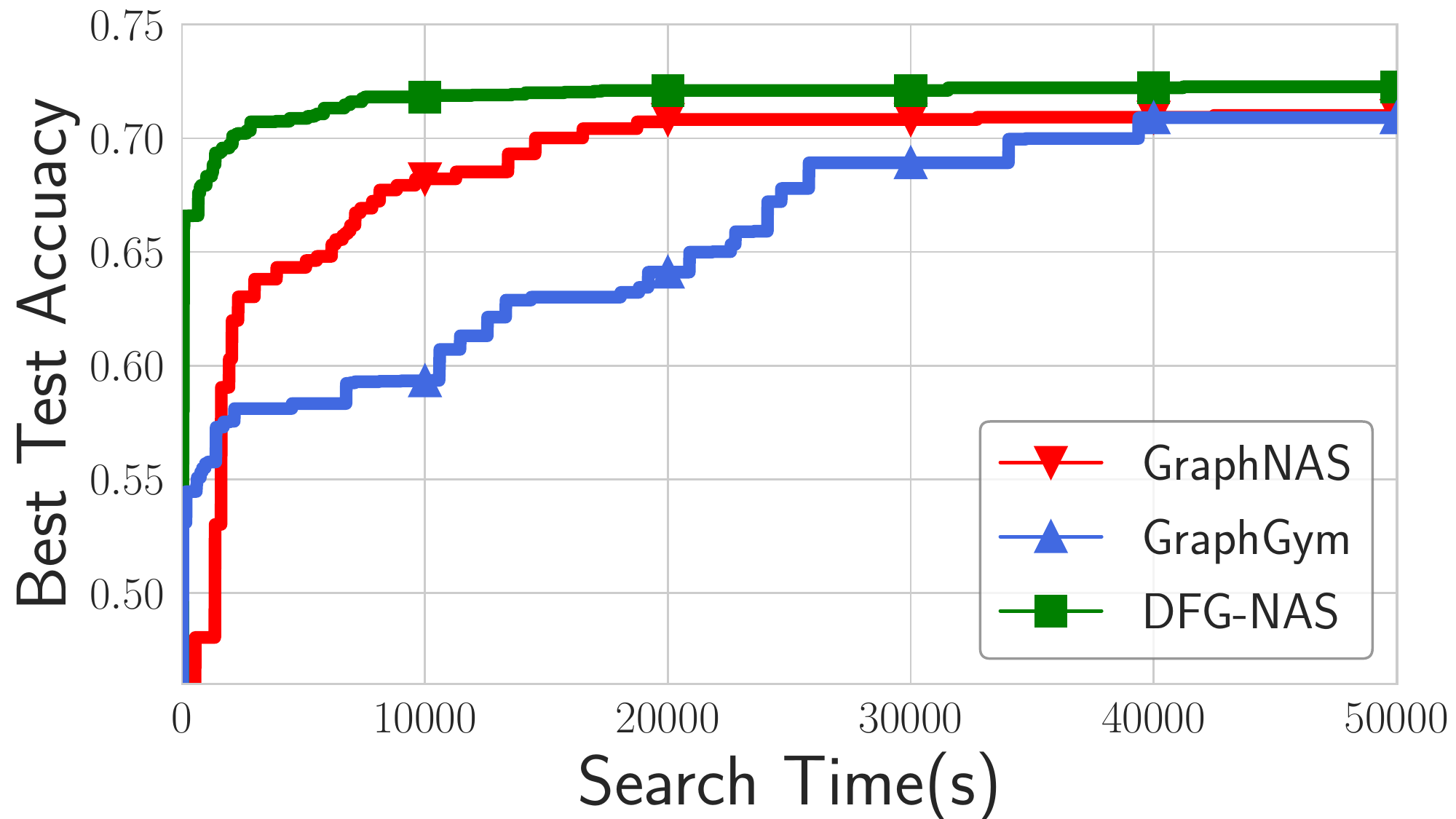}}
	}
% 	\vspace{-3mm}
	\caption{Test results during neural architecture search on four datasets.}
	\label{fig:nas}
% 	\vspace{-3mm}
\end{figure*}

\begin{table}[tpb!]
% \vspace{-5mm}
\caption{Ablation study on Gate operations.
% \zwt{1)在search space里去掉gate，比较smoothness说明没有gate会over-smooth 2）搜索结束后去掉gate，说明如果没有gate的话，模型更趋向于浅层，因为能避免oversmoothing，这样就做不深，无法有效获取信息}
}
% \vspace{-2mm}
\centering
{
\noindent
\renewcommand{\multirowsetup}{\centering}
\resizebox{0.95\linewidth}{!}{
\begin{tabular}{cccccc}
\toprule
\textbf{Methods} & \textbf{Cora} & \textbf{Citeseer} & \textbf{PubMed} & \textbf{ogbn-arxiv}\\
\hline
A w/o Gate & 82.0$\pm$0.3 & 70.8$\pm$0.5 &78.1$\pm$0.1 & 68.5$\pm$0.6\\
S w/o Gate & 83.6$\pm$0.2 & 72.6$\pm$0.2 & 79.3$\pm$0.3 & 71.5$\pm$0.4\\
\hline
\sys & \textbf{85.2$\pm$0.2} & \textbf{74.1$\pm$0.4} & \textbf{81.1$\pm$0.3} & \textbf{72.3$\pm$0.2}\\
\bottomrule
\end{tabular}}}
\label{withoutgate}
\vspace{-3mm}
\end{table}

\subsection{Comparison with NAS methods}
Figure~\ref{fig:nas} and Table~\ref{table:performance} demonstrate the performance of \sys compared with existing NAS methods for GNN. 
Since AutoGNN is not open-sourced, we only report the results on three datasets from its paper.
From Figure~\ref{fig:nas}, we observe that \sys consistently outperforms the compared NAS methods.
The main reason is that the existing NAS methods only consider a fixed pipeline pattern of propagation and transformation operations in their architecture design, which may not be the optimal order as introduced in Section~\ref{sec:number_order}.
When the search budget exhausts, \sys outperforms the second-best baseline by a margin of 0.3-1.5\% on test accuracy.
Compared with the second-best baseline GraphNAS, \sys achieves 6.97-15.96x speedups when achieving the same test accuracy on four datasets, which indicates the superior efficiency of our proposed method.

% \red{Remarkably, it takes \sys less than 100 iterations to achieve the same results of GraphNAS with 500 iterations on all four datasets. 
% After 500 iterations, \sys outperforms the second-best baseline by a margin of 0.3-1.5\% on test accuracy. }

% \zwt{比GraphNAS更快搜出来好结果}

\subsection{Ablation Study}
In this part, we present an ablation study to show the effectiveness of three designs in \sys, which are 1) the Gate operation; 2) the disentanglement of architectures and 3) the skip-connection operation. 
% that \sys can deal with the two limitations mentioned in Section~\ref{sec:introduction}, which are 1) Risk of Over-smoothing and 2) Lack of Flexibility.

In Table~\ref{withoutgate}, we compare \sys with two baselines: 1) A w/o Gate: the resulting architecture with the Gate operation disabled and 2) S w/o Gate: the searched results over a reduced space without the Gate operation in architecture design. 
We observe that the accuracy declines on all datasets. 
Remarkably, the accuracy drops by 3.0-3.8\% for architectures without Gate. 
The reason is that without the Gate operation, the transformation steps only take the last output of the propagation steps as inputs, which may suffer from a risk of over-smoothing. 
The Gate operation dynamically aggregates the information from all propagation steps and thus controls the smoothness of different nodes in a node-adaptive way.
In addition, when the Gate operation is removed from the architecture space, the accuracy drops by 0.8-1.8\%. 
In this setting, the searched architectures tend to have shallow propagation (i.e., only 3-5 steps) on all datasets, which is not enough to capture sufficient neighborhood information for each node. These ablation results demonstrate the importance of the gating mechanism on GNN architecture design.

\begin{table}[tpb!]
% \vspace{-5mm}
\caption{Ablation study on pipeline pattern.}
% \vspace{-2mm}
\centering
{
\noindent
\renewcommand{\multirowsetup}{\centering}
\resizebox{0.95\linewidth}{!}{
\begin{tabular}{cccccc}
\toprule
\textbf{Methods} & \textbf{Cora} & \textbf{Citeseer} & \textbf{PubMed} & \textbf{ogbn-arxiv}\\
\hline
\textbf{\texttt{P}} before \textbf{\texttt{T}} & 83.2$\pm$0.3 & 72.7$\pm$0.2 &80.0$\pm$0.4 & 69.4$\pm$0.1\\
\textbf{\texttt{T}} before \textbf{\texttt{P}} & 83.6$\pm$0.5 & 71.8$\pm$0.3 &79.0$\pm$0.5 & 70.0$\pm$0.1\\
Alternate \textbf{\texttt{P}} and \textbf{\texttt{T}} & 82.5$\pm$0.4 & 69.9$\pm$0.6 &78.1$\pm$0.3 & 68.1$\pm$0.4\\
\hline 
\sys & \textbf{85.2$\pm$0.2} & \textbf{74.1$\pm$0.4} & \textbf{81.1$\pm$0.3} & \textbf{72.3$\pm$0.2} \\
\bottomrule
\end{tabular}}}
\label{designspace}
% \vspace{-3mm}
\end{table}

\begin{table}[h]
% \vspace{-5mm}
\caption{Ablation study on skip-connection.
}
% \vspace{-2mm}
\centering
{
\noindent
\renewcommand{\multirowsetup}{\centering}
\resizebox{0.95\linewidth}{!}{
\begin{tabular}{cccccc}
\toprule
\textbf{Methods} & \textbf{Cora} & \textbf{Citeseer} & \textbf{PubMed} \\
\hline
A w/o Skip & 83.7$\pm$0.5 & 71.6$\pm$0.5 &81.0$\pm$0.3 \\
S w/o Skip & 83.9$\pm$0.4 & 72.8$\pm$0.6 & 80.6$\pm$0.2 \\
\hline
\sys & \textbf{85.2$\pm$0.2} & \textbf{74.1$\pm$0.4} & \textbf{81.1$\pm$0.3}\\
\bottomrule
\end{tabular}}
}
\label{withoutskip}
% \vspace{-3mm}
\end{table}

To show the flexibility of the search space in \sys, we consider the following search spaces inspired by existing designs of propagation (P) and transformation (T): (1) \textbf{\texttt{P}} before \textbf{\texttt{T}}, (2) \textbf{\texttt{T}} before \textbf{\texttt{P}}, and (3) alternate \textbf{\texttt{P}} and \textbf{\texttt{T}}. 
The depths of \textbf{\texttt{P}} and \textbf{\texttt{T}} range from 1 to 10, which leads to 100 unique architectures in (1) and (2), and ten unique architectures in (3). 
We exhaustively evaluate all possible architectures in the baseline search space, and Table~\ref{designspace} shows the best observed performance over each search space. 
Among the baselines, alternate \textbf{\texttt{P}} and \textbf{\texttt{T}} has the worst performance, which is consistent with the observations in previous studies~\cite{liu2020towards}. 
In addition, none of the three baselines dominate the others on all datasets, which indicates the necessity of a more flexible search on macro-architecture.
Since the search space of \sys does not restrict the pipeline pattern of \textbf{\texttt{P}} and \textbf{\texttt{T}}, the space of \sys covers the compared baselines, and the architectures are more flexible than those from the baselines. 
As a result, \sys achieves an increase of 1.1\% to 2.3\% on all datasets.
% \zwt{不flexibility的：呼应Section 3, flexible PT设计的好处，对比1-20层的下列方法：1）先P后T的SGC类 2）PT耦合的GCN类 3）先T后P的APPNP类。P的个数是1到10，T的个数也是1到10，这样对SGC和APPNP就有100种组合。GCN层数就是从2，4,6,8,...,20层。在三个citation network上证明，这些方法搜出来的效果都没有我们的好}

To show the effectiveness of skip-connection. In Table~\ref{withoutskip}, we compare \sys  with two baselines: 1) A w/o Skip: the resulting architecture with the Skip-connection disabled and 2) S w/o Skip: the searched results over a reduced space without the skip-connection in architecture design. 
We observe that the accuracy declines on all datasets. 
And we also find when the skip-connection is removed from the search space, the architectures searched by our method will have less T operations.

To summarize, the searched architecture performance is improved by introducing the Gate operation in propagation steps, disentangling \textbf{\texttt{P}} and \textbf{\texttt{T}} in architecture designs, and the skip-connection in \textbf{\texttt{T}} operations. All of them may help unleash the full potential of GNNs and inspire more powerful manual designs.

\subsection{Interpretability}
% \zwt{@ 先放着，让杨老师来看
% 1. 观察已经搜到的结果，看纵坐标Smoothing Level随着横坐标Operator操作（PT）次数的变化，重点观察最终的Smoothing Level值和accuracy的排序的相关性，统计第top 1，top 10，20，30个，另外把DAGNN也加上}

% \zwt{2. 追踪遗传算法搜到的网络结构的平均smoothing level，横轴是遗传算法的迭代轮数（比如200论）， 纵轴对应网络的平均的smoothing level，期待效果：上升饱和}

\begin{figure}[tpb]
    \centering
    \subfigure[Cora]{
    \label{fig:evo_cora}
		% Requires \usepackage{graphicx}
		\scalebox{0.475}{
    \includegraphics[width=0.98\linewidth]{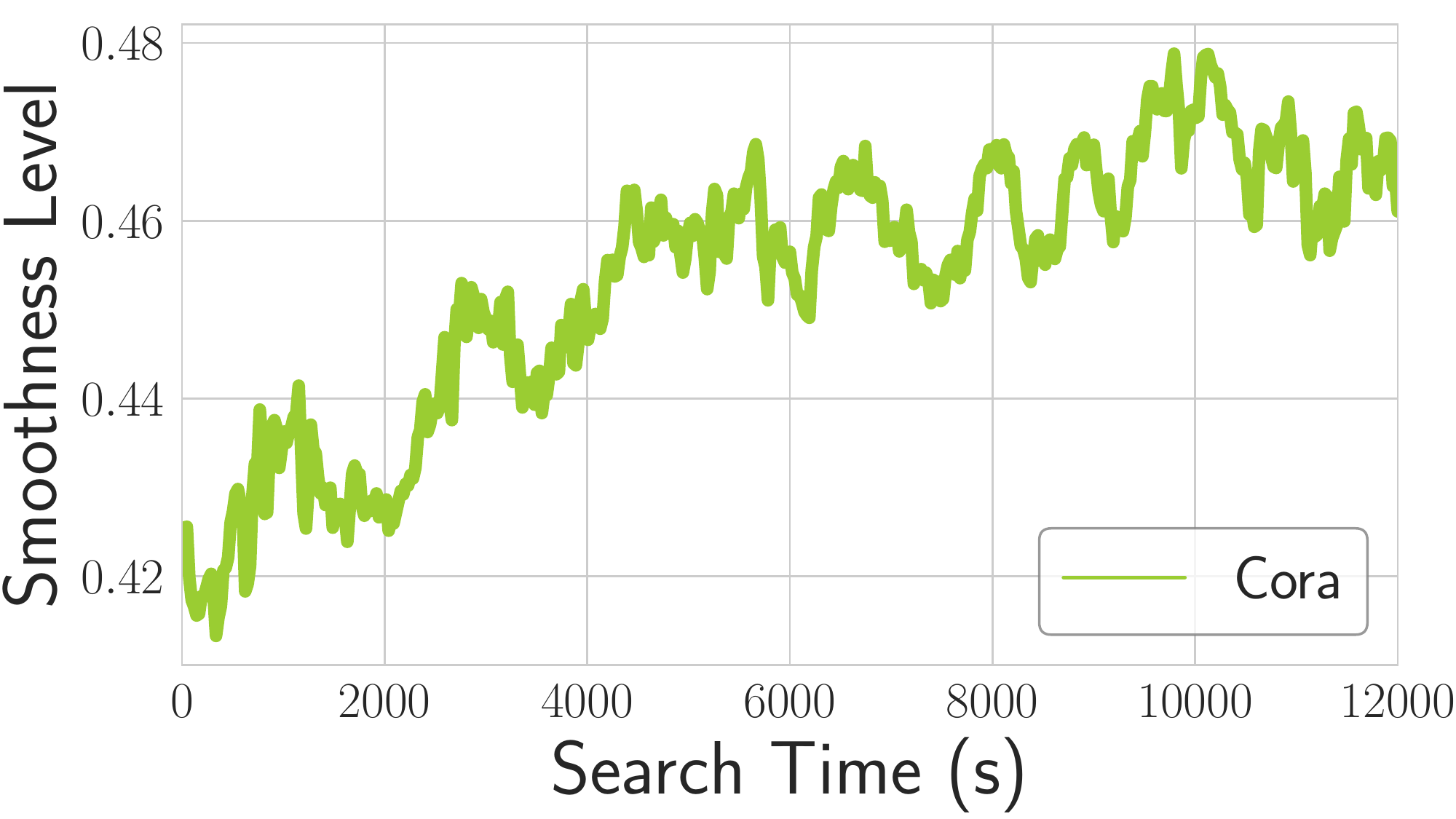}}
	}
    \subfigure[Pubmed]{
    \label{fig:evo_pubmed}
		% Requires \usepackage{graphicx}
		\scalebox{0.475}{
    \includegraphics[width=0.98\linewidth]{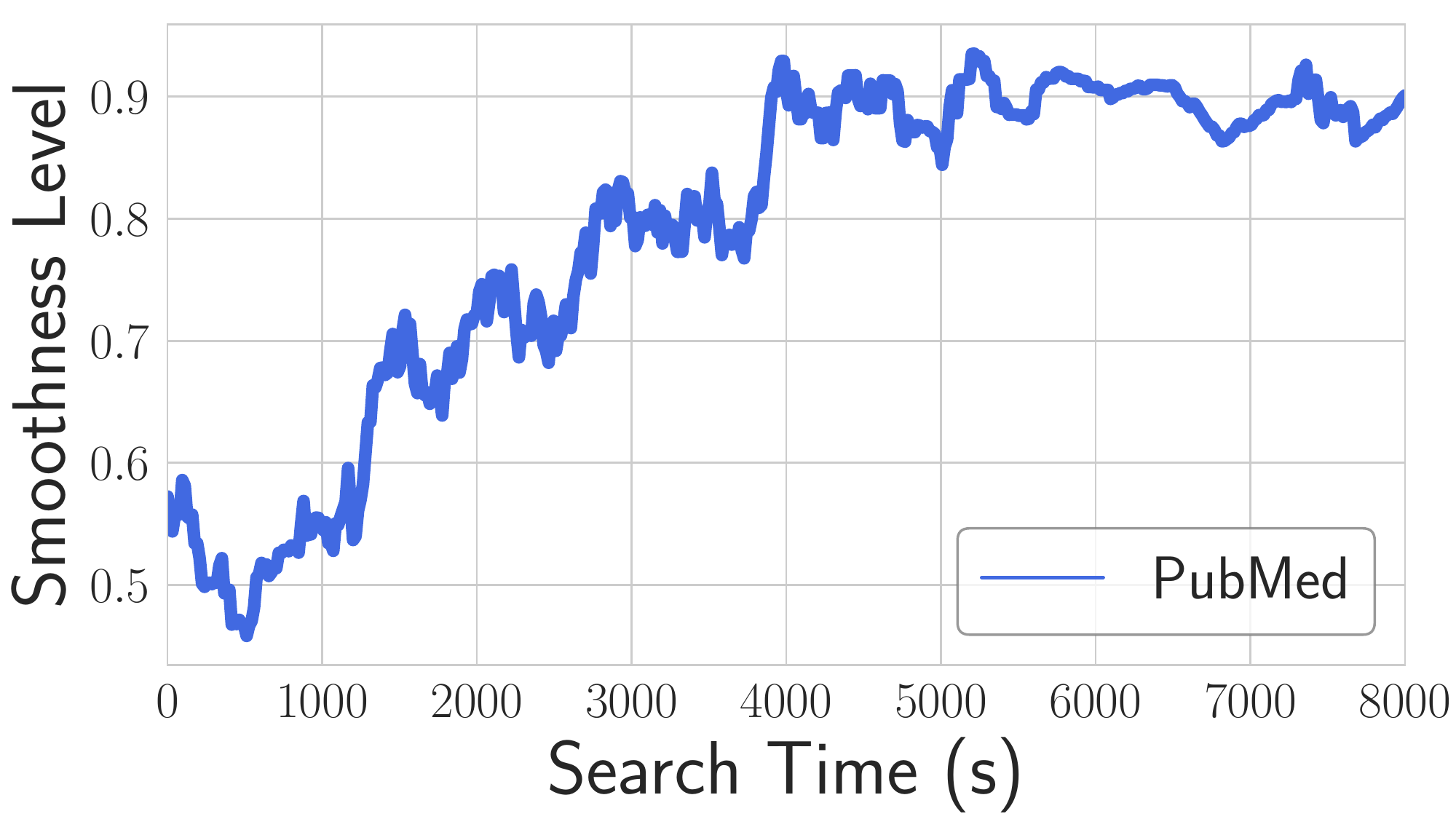}}
	}
    % \vspace{-3mm}
    \caption{Average smoothness over iterations on two datasets.}
    \label{fig:trend}
    % \vspace{-2mm}
\end{figure}

In Figure~\ref{fig:trend}, we plot the average smoothness of the last layer outputs of architectures in the population during each iteration on two datasets. 
The definition of smoothness is provided in Equation~\ref{equation:smoothness}.
Though graph datasets differ in the requirement of smoothness, both two curves are generally saturating. 
In the first half of the search process, the smoothness quickly increases due to the frequent elimination of bad initial individuals. 
Then, the smoothness fluctuates around a certain level (0.46 for Cora and 0.90 for Pubmed), which implies that the search algorithm in \sys can ensure sufficient smoothness of inputs in the searched architectures, and meanwhile avoid over-smoothing. 

\begin{figure}[tpb]
    \centering
    \subfigure[Trend of \textbf{\texttt{P}} on sparsity]{
    \label{fig:trend_p}
		% Requires \usepackage{graphicx}
		\scalebox{0.475}{
    \includegraphics[width=0.98\linewidth]{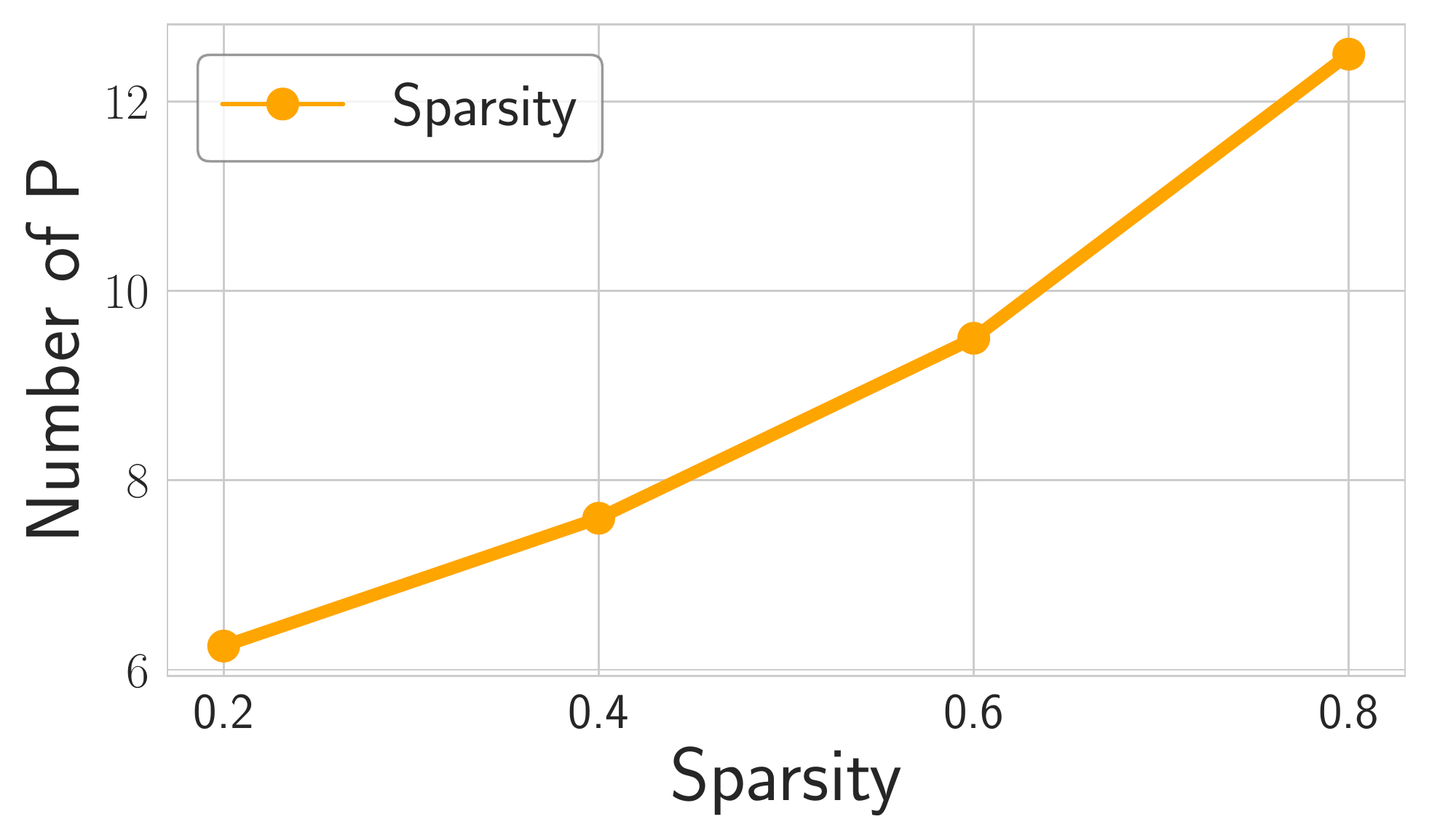}}
	}
    \subfigure[Trend of \textbf{\texttt{T}} on data size]{
    \label{fig:trend_t}
		% Requires \usepackage{graphicx}
		\scalebox{0.475}{
    \includegraphics[width=0.98\linewidth]{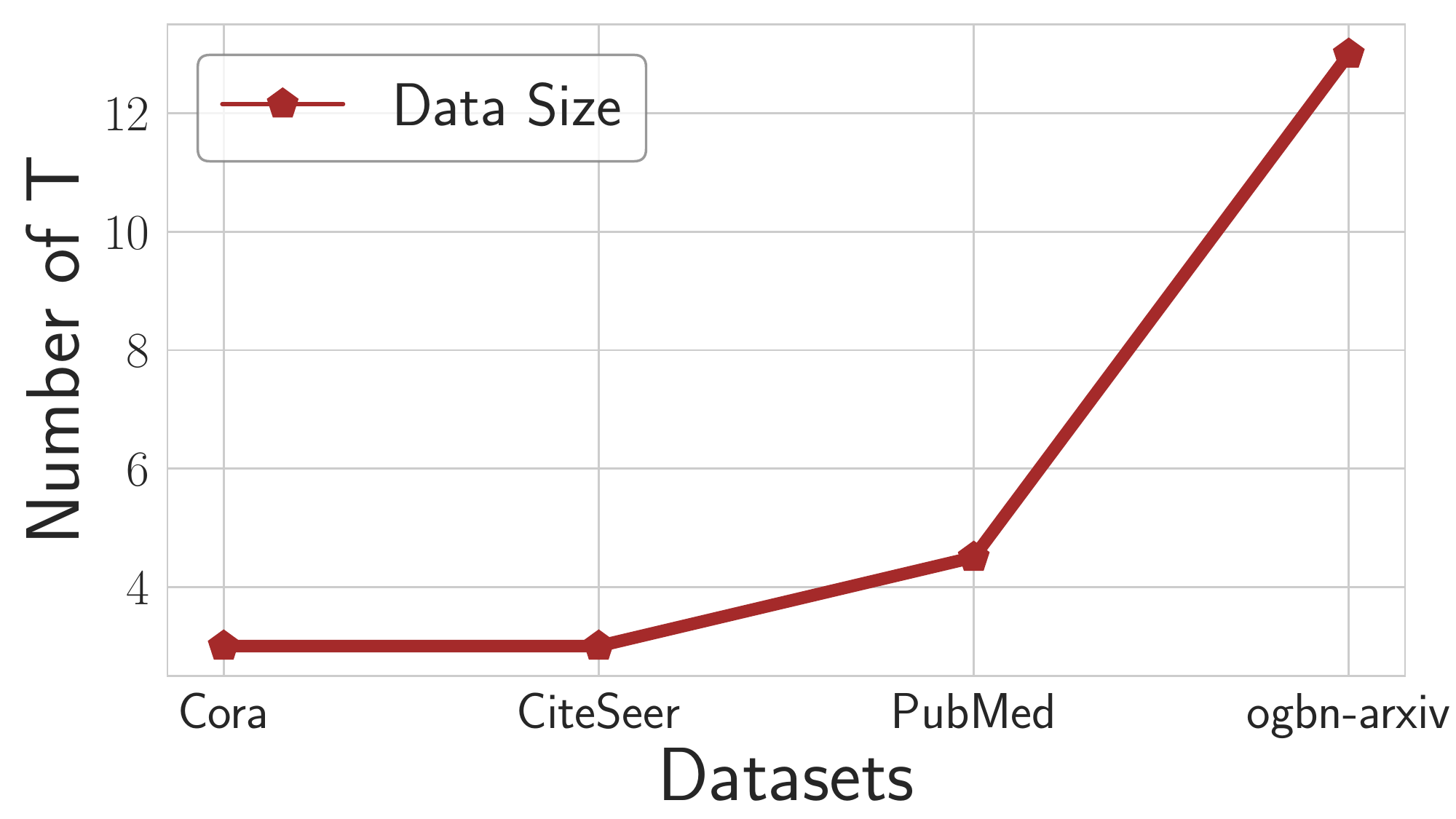}}
	}
    % \vspace{-3mm}
    \caption{Left: Average
     number of \textbf{\texttt{P}} operations along with the increased sparsity on PubMed. Right: Average number of \textbf{\texttt{T}} operations along with the increased nodes of different datasets.}
    \label{interpretability}
    % \vspace{-2mm}
\end{figure}

Finally, we study how the sparsity and size of datasets influence the number of \textbf{\texttt{P}} and \textbf{\texttt{T}} in the searched architectures, respectively. 
To simulate datasets with different sparsity, we randomly delete some of the edges in PubMed. 
Figure~\ref{fig:trend_p} shows that the average number of \textbf{\texttt{P}} in top-10 architectures increases when the dataset grows sparser. 
The reason is that, when the graph is sparse, the architecture should contain more propagation steps so that each node is able to capture sufficient neighborhood information for classification. 
On the other hand, when the size of datasets grows larger, the architecture should include more transformation steps to ensure the expressive power. 
As shown in Figure~\ref{fig:trend_t}, the average number of \textbf{\texttt{T}} is similar in Core and Citeseer, which includes about 3k samples. 
The number further increases to 4.5 and 13.0 on larger datasets PubMed and ogbn-arxiv with 20k and 169k samples, respectively.
The trend of \textbf{\texttt{P}} and \textbf{\texttt{T}} discovered by \sys may help guide the manual design of flexible GNN architectures, i.e., more propagation steps are required when the graph is sparse, while more transformation steps are needed for the large graph.
% \zwt{3. PubMed数据集上取sparse情况下（删掉一些边）top 10的网络结构，横轴是边的稀疏程度，纵轴是这些top 10网络结构的1）P的平均次数随稀疏度变化 2）T的平均次数随样本量变化 }

% \zwt{画图放到附录里，前面说了我们的方法可以Deep and Flexible，这里要显示一下}

\section{Conclusion}
In this paper, we proposed \sys, a new method that allows both flexible and deep graph neural architecture search. 
The key idea of \sys is to dis-aggregate the design pipeline of GNN generation, allowing a flexible permutations and combinations of the two basic operations in GNNs: propagation and transformation. 
In addition, we also analyzed how both operations influence the smoothness. 
Based on the observation, we adopted the gating mechanism and skip-connection mechanisms to support very deep GNN pipelines.
We provided the genetic algorithm to search
for a good permutations and combinations in GNNs.
% We conducted extensive experiments to verify the effectiveness of \sys 
The experimental results on four different graph datasets showed that \sys achieves an accuracy improvement of up to 0.9\% over state-of-the-art manual designs and brings up to 15.96x speedups over existing G-NAS methods.

% \zwt{补充实验结果：cora 比mannual 高0.9
% 比NAS高1.5；Citeseer 0.8
% NAS 0.3；PubMed都是 0.6
% 0.6；OGB 0.4 manual
% NAS 0.6}.

\section*{Acknowledgement}
This work is supported by NSFC (No. 61832001, 61972004), and Alibaba Group through Alibaba Innovative Research Program. 

\bibliography{reference}
\bibliographystyle{icml2022}
\clearpage

\appendix
\section{Outline}
The appendix is organized as follows:
\begin{description}
    \item[A.1] Datasets Description.
    \item[A.2] Hyper-parameters Setting.
    \item[A.3] Compared Baselines.
    \item[A.4] Efficiency Analysis.
    \item[A.5] Reproduction Instructions.
    \item[A.6] The Best Architecture Searched by \sys.
\end{description}

\subsection{Dataset Description}
\label{dataset_description}
\textbf{Cora}, \textbf{Citeseer}, and \textbf{Pubmed}\footnote{https://github.com/tkipf/gcn/tree/master/gcn/data} are three popular citation network datasets, and we follow the public training/validation/test split in GCN ~\cite{DBLP:conf/iclr/KipfW17}.
In these three networks, papers from different topics are considered as nodes, and the edges are citations among the papers. The node attributes are binary word vectors, and class labels are the topics papers belong to.

\noindent\textbf{ogbn-arxiv} is a directed graph, representing the citation network among all Computer Science (CS) arXiv papers indexed by MAG. The training/validation/test split in our experiment is the same as the public version. The public version provided by OGB\footnote{https://ogb.stanford.edu/docs/nodeprop/\#ogbn-arxiv}is used in our paper.

\subsection{Hyper-parameters Setting}
\label{hpo}
For the architecture search, the number of the population set $k$ and the maximum generation times $T$ in Algorithm~\ref{alg:search} are 20 and 500 for all datasets.
For the training of GNN architectures, we follow the same hyper-parameter in their original paper and tune it with OpenBox~\cite{li2021openbox}.
The training budget of each searched GNN architecture in \sys is 200 epochs for three citation networks and 500 epochs for the ogbn-arxiv dataset.
Specifically, we train them using Adam optimizer with a learning rate of 0.02 for Cora, 0.03 for Citeseer, 0.1 for PubMed, and 0.001 for ogbn-arxiv. The regularization factor is 5e-4 for all datasets. 
We apply dropout to all feature vectors with rates of 0.5 for Cora and Citeseer, and 0.3 for PubMed and ogbn-arxiv.
Besides, the dropout between different GNN layers is 0.8 for Cora and Citeseer, and 0.5 for PubMed and ogbn-arxiv. 
At last, the hidden size of each GNN layer is 128 for Cora and ogbn-arxiv, 256 for Citeseer, and 512 for ogbn-arxiv. 

\begin{table}[tp]
\small
\centering
\caption{Overview of the Four Datasets.} \label{Dataset}
\resizebox{0.96\linewidth}{!}{
\begin{tabular}{ccccccc}
\toprule
\textbf{Dataset}&\textbf{\#Nodes}& \textbf{\#Features}&\textbf{\#Edges}&\textbf{\#Classes}&\textbf{\#Train/Val/Test}\\
\midrule
Cora& 2,708 & 1,433 &5,429&7& 1,208/500/1,000 \\
Citeseer& 3,327 & 3,703&4,732&6& 1,827/500/1,000 \\
Pubmed& 19,717 & 500 &44,338&3& 18,217/500/1,000\\
ogbn-arxiv& 169,343 & 128 & 1,166,243 & 40 & 90,941/29,799/48,603 \\
\bottomrule
\end{tabular}}
\label{table:data}
\end{table}

\begin{table*}[tp]
\caption{The searched GNN architecture in different datasets.}
% \vspace{-5mm}
% \vspace{-2mm}
\centering
{
\noindent
\renewcommand{\multirowsetup}{\centering}
\resizebox{0.9\linewidth}{!}{
\begin{tabular}{cc}
\toprule
\textbf{Datasets} & \textbf{Searched Architectures (in sequence)}  \\
\midrule 
Cora& [\textbf{\texttt{T}}, \textbf{\texttt{P}}, \textbf{\texttt{P}}, \textbf{\texttt{P}}, \textbf{\texttt{P}}, \textbf{\texttt{P}}, \textbf{\texttt{T}}, \textbf{\texttt{P}}, \textbf{\texttt{P}}, \textbf{\texttt{T}}]\\
Citeseer& [\textbf{\texttt{T}}, \textbf{\texttt{P}}, \textbf{\texttt{P}}, \textbf{\texttt{P}}, \textbf{\texttt{P}}, \textbf{\texttt{P}}, \textbf{\texttt{P}}, \textbf{\texttt{P}}, \textbf{\texttt{P}}, \textbf{\texttt{P}}, \textbf{\texttt{P}}, \textbf{\texttt{P}}, \textbf{\texttt{T}}, \textbf{\texttt{P}}, \textbf{\texttt{T}}]\\
\textbf{\texttt{P}}ubMed& [\textbf{\texttt{T}}, \textbf{\texttt{P}}, \textbf{\texttt{P}}, \textbf{\texttt{P}}, \textbf{\texttt{P}}, \textbf{\texttt{P}}, \textbf{\texttt{P}}, \textbf{\texttt{P}}, \textbf{\texttt{P}}, \textbf{\texttt{P}}, \textbf{\texttt{P}}, \textbf{\texttt{P}}, \textbf{\texttt{P}}, \textbf{\texttt{P}}, \textbf{\texttt{P}}, \textbf{\texttt{P}}, \textbf{\texttt{P}}, \textbf{\texttt{P}}, \textbf{\texttt{P}}, \textbf{\texttt{P}}, \textbf{\texttt{P}}, \textbf{\texttt{P}}, \textbf{\texttt{P}}, \textbf{\texttt{T}}] \\
ogbn-arxiv& [\textbf{\texttt{T}}, \textbf{\texttt{P}}, \textbf{\texttt{P}}, \textbf{\texttt{P}}, \textbf{\texttt{P}}, \textbf{\texttt{P}}, \textbf{\texttt{P}}, \textbf{\texttt{P}}, \textbf{\texttt{P}}, \textbf{\texttt{P}}, \textbf{\texttt{P}}, \textbf{\texttt{T}}, \textbf{\texttt{T}}, \textbf{\texttt{P}}, \textbf{\texttt{P}}, \textbf{\texttt{P}}, \textbf{\texttt{T}}, \textbf{\texttt{P}}, \textbf{\texttt{T}}, \textbf{\texttt{P}}, \textbf{\texttt{P}}, \textbf{\texttt{T}}, \textbf{\texttt{T}}, \textbf{\texttt{P}}, \textbf{\texttt{P}}, \textbf{\texttt{T}}, \textbf{\texttt{T}}, \textbf{\texttt{T}}, \textbf{\texttt{P}}, \textbf{\texttt{T}}, \textbf{\texttt{P}}, \textbf{\texttt{P}}, \textbf{\texttt{P}}, \textbf{\texttt{T}}, \textbf{\texttt{T}}, \textbf{\texttt{P}}, \textbf{\texttt{T}}, \textbf{\texttt{P}}, \textbf{\texttt{T}}, \textbf{\texttt{P}}, \textbf{\texttt{P}}, \textbf{\texttt{T}}, \textbf{\texttt{P}}, \textbf{\texttt{T}}, \textbf{\texttt{T}}]\\
\bottomrule
\end{tabular}}}
\label{Searched-results}
% \vspace{-3mm}
\end{table*}

\subsection{Compared Baselines.}
\label{baseline_methods}
For existing manual GNNs, we summarize the existing baselines according to the pipeline pattern of propagation ($\textbf{\texttt{P}}$) and transformation ($\textbf{\texttt{T}}$) operations. Specifically, they can be classified into the following three types.

(1) Alternate $\textbf{\texttt{P}}$ and $\textbf{\texttt{T}}$:
\begin{itemize}
    \item Graph Convolutional Network(GCN) ~\cite{DBLP:conf/iclr/KipfW17}: GCN adopts an efficient layer-wise propagation rule that is based on a first-order approx- imation of spectral convolutions on graphs.
    \item Graph Attention Networks(GAT) ~\cite{velivckovic2017graph}: GAT leverages masked self-attention layers to specify different weights to different nodes in a neighborhood, thus better represent graph information.
    \item GraphSAGE ~\cite{hamilton2017inductive}: GraphSAGE is an inductive framework that leverages node attribute information to efficiently generate representations on previously unseen data.
\end{itemize}

(2) $\textbf{\texttt{P}}$ before $\textbf{\texttt{T}}$:
\begin{itemize}
    \item Simplified GCN (SGC) ~\cite{wu2019simplifying}: SGC simplifies GCN by removing nonlinearities and collapsing weight matrices between consecutive layers. 
    \item Scalable Inception Graph Neural Networks (SIGN) ~\cite{rossi2020sign}: SIGN is an efficient and scalable graph embedding method that sidesteps graph sampling in GCN and uses different local graph operators to support different tasks. 
     \item  Simple Spectral Graph Convolution (S$^2$GC)~\cite{zhu2021simple}: S$^2$GC is a trade-off of low- and high-pass filter bands which capture the global and local contexts of each node
\end{itemize}

(3) $\textbf{\texttt{T}}$ before $\textbf{\texttt{P}}$
\begin{itemize}
    \item APPNP ~\cite{DBLP:conf/iclr/KlicperaBG19}: APPNP has the ability to use the relationship between graph convolution networks (GCN) and PageRank to derive improved node representations.
    \item Adaptive Propagation Graph Convolution Network (AP-GCN) ~\cite{spinelli2020adaptive}: AP-GCN uses a halting unit to decide a receptive range of a given node. 
    \item Deep Adaptive Graph Neural Network (DAGNN):  DAGNN decouples the propagation and transformation operations, and it can adaptively incorporate information from large receptive fields.
\end{itemize}

For G-NAS methods, the compared baselines are as follows:
\begin{itemize}
    \item Auto-GNN~\cite{zhou2019auto}: Auto-GNN is a reinforced conservative search strategy by adopting both RNNs and evolutionary algorithms in the controller.
    \item GraphNAS~\cite{gao2019graphnas}: GraphNAS is a reinforcement learning-based method that uses an RNN controller to sample from the multiple architectures sequentially.
    \item GraphGym~\cite{you2020design}: GraphGym is a variant of random search on a general GNN search space that considers intra-layer design, inter-layer design, and training configurations.
\end{itemize}
Note that we implement GraphNAS~\footnote{https://github.com/GraphNAS/GraphNAS} and GraphGym~\footnote{https://github.com/snap-stanford/GraphGym} according to their open-sourced version. 
 
\subsection{ Efficiency Analysis} 
The time complexity of the EA method is $O(m)$, where $m$ is the population size. In other words, the method is independent of the number of evaluations, and runs as fast as random search.
While $\textbf{\texttt{P}}$ or $\textbf{\texttt{T}}$ can be added infinitely, previous methods for finite spaces can not be directly applied to our design space. 
And thus we perform the end-to-end comparisons in our paper. 
Figure~\ref{fig:nas} shows that \sys achieves similar performance to other methods with less search time. 
As our searching algorithm is relatively simple, we attribute this gain to the design of our search space, i.e., the well-designed architecture in our search space outperforms the precious ones. 

\subsection{ Reproduction Instructions}
\label{eno}
The experiments are conducted on a machine with Intel(R) Xeon(R) Gold 5120 CPU @ 2.20GHz, and a single NVIDIA TITAN RTX GPU with 24GB GPU memory.
The operating system of the machine is Ubuntu 16.04.
For software versions, we use Python 3.6, Pytorch 1.7.1, and CUDA 10.1.
Our code is available in the anonymized repository \url{https://github.com/PKU-DAIR/DFG-NAS}.

\subsection{The Best Architecture Searched by \sys}
\label{sec:searched_results}
The best GNN architecture searched by \sys on different graph datasets is summarized in Table~\ref{Searched-results}. Note that each GNN architecture will begin with a $\textbf{\texttt{T}}$ operation for dimension reduction and end with a $\textbf{\texttt{T}}$ operation for getting the softmax outputs.

% \zwt{TODO}
% \zwt{1.Combination order有这种说法吗？组合顺序？一般是排列顺序：全文替换为Pipeline Pattern}

% \zwt{2. Pipeline pattern听着奇怪}

% \zwt{3.Graph neural architecture search 和GNN Architecture Search统一写法: 都改成G-NAS}
% \begin{figure*}[!t]
% \begin{minipage}{\textwidth}
% \Large
% \centering
% \medskip
% \textbf{Supplementary Materials for \sys}
% \medskip
% \end{minipage}
% \end{figure*}
% \newpage
\end{document}